\documentclass[10pt,twocolumn,letterpaper]{article}
\usepackage{amssymb}
\usepackage{mathtools}

\usepackage{booktabs}
\usepackage{multirow}
\usepackage{colortbl}
\usepackage[pagenumbers]{cvpr} % To force page numbers, e.g. for an arXiv version

\definecolor{cvprblue}{rgb}{0.21,0.49,0.74}
\usepackage[pagebackref,breaklinks,colorlinks,citecolor=cvprblue]{hyperref}

\def\paperID{12018} % *** Enter the Paper ID here
\def\confName{CVPR}
\def\confYear{2025}

\title{Pinpoint Counterfactuals: Reducing social bias in foundation models via localized counterfactual generation.}

\author{Kirill Sirotkin\\
VPU Lab\\
UAM, Madrid\thanks{Universidad Autónoma de Madrid, Departamento TEC, Video Processing and Understanding Lab}\\
{\tt\small kirill.sirotkin@uam.es}
\and
Marcos Escudero-Viñolo\\
VPU Lab\\
UAM, Madrid\footnotemark[1]\\
{\tt\small marcos.escudero@uam.es}
\and
Pablo Carballeira\\
VPU Lab\\
UAM, Madrid\footnotemark[1]\\
{\tt\small pablo.carballeira@uam.es}
\and
Mayug Maniparambil\thanks{{\tt\small mayug.maniparambil2@mail.dcu.ie}}
\\
ML Labs\\
DCU, Dublin
\and
Catarina Barata\thanks{{\tt\small ana.c.fidalgo.barata@tecnico.ulisboa.pt}}\\
LARSyS\\
LRS, Lisbon
\and
Noel E. O’Connor\\
ML Labs\\
DCU, Dublin\\
{\tt\small Noel.OConnor@dcu.ie}}
\begin{document}
\maketitle
\begin{abstract}
Foundation models trained on web-scraped datasets propagate societal biases to downstream tasks. While counterfactual generation enables bias analysis, existing methods introduce artifacts by modifying contextual elements like clothing and background. We present a localized counterfactual generation method that preserves image context by constraining counterfactual modifications to specific attribute-relevant regions through automated masking and guided inpainting. When applied to the Conceptual Captions dataset for creating gender counterfactuals, our method results in higher visual and semantic fidelity than state-of-the-art alternatives, while maintaining the performance of models trained using only real data on non-human-centric tasks. Models fine-tuned with our counterfactuals demonstrate measurable bias reduction across multiple metrics, including a decrease in gender classification disparity and balanced person preference scores, while preserving ImageNet zero-shot performance. The results establish a framework for creating balanced datasets that enable both accurate bias profiling and effective mitigation.
\end{abstract}    
\section{Introduction}
\label{sec:intro}
\begin{figure}[ht]
  \centering
  
 % \begin{subfigure}[l]{0.15\textwidth}
 %    \centering
 %      \includegraphics[height=2.3cm]{figures/main/banana_woman.jpg}
 %     \caption{Unrealistic \cite{bib_coco_counterfactuals}}
 %     \label{fig_banana}
 % \end{subfigure}
\begin{subfigure}[l]{0.11\textwidth}
\centering
      \includegraphics[height=2cm]{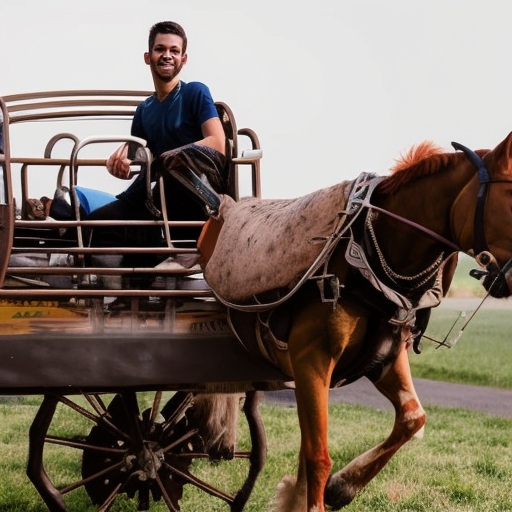}
     % \caption{Unrealistic \cite{bib_coco_counterfactuals}}
     \label{fig_horse}
 \end{subfigure}
 \begin{subfigure}[l]{0.11\textwidth}
 \centering
      \includegraphics[height=2cm]{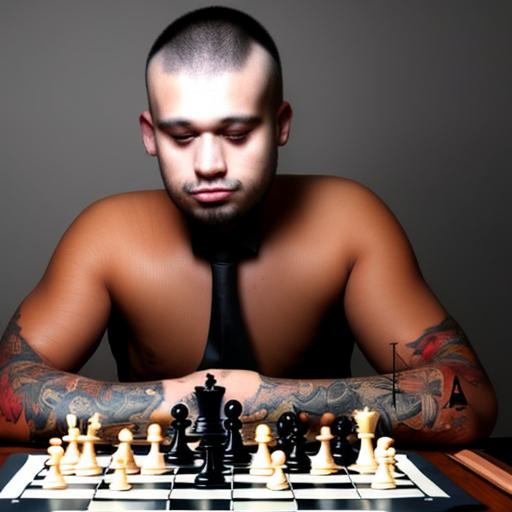}
     % \caption{Shirtless man \cite{bib_social_counterfactuals}}
     \label{fig_gm}
 \end{subfigure}
 % \begin{subfigure}[l]{0.15\textwidth}
 % \centering
 %      \includegraphics[height=1.5cm]{figures/main/cleaner.jpg}
 %     \caption{Shirtless man \cite{bib_social_counterfactuals}}
 %     \label{fig_cleaner}
 % \end{subfigure}
 \begin{subfigure}[l]{0.11\textwidth}
 \centering
      \includegraphics[height=2cm]{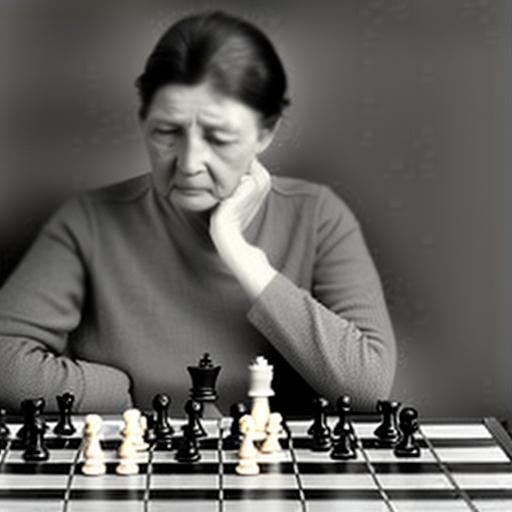}
     % \caption{Gray photo \cite{bib_social_counterfactuals}}
     \label{fig_gray}
 \end{subfigure}
  \begin{subfigure}[l]{0.11\textwidth}
 \centering
      \includegraphics[height=2cm]{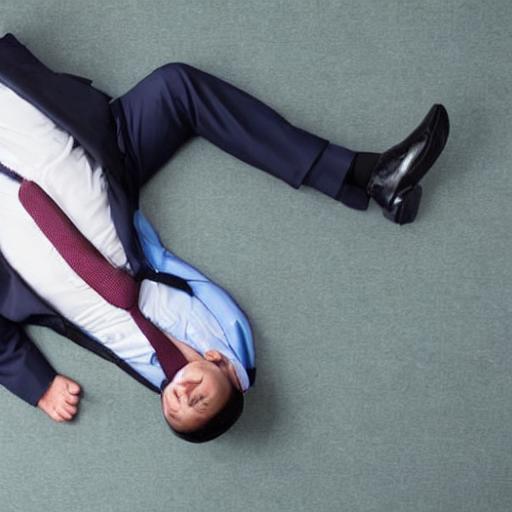}
     % \caption{Unrealistic \cite{bib_future_bias}}
     \label{fig_legs}
 \end{subfigure}
  \begin{subfigure}[l]{0.11\textwidth}
 \centering
      \includegraphics[height=3cm]{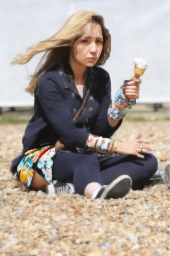}
     % \caption{Unrealistic \cite{bib_future_bias}}
 \end{subfigure}
  \begin{subfigure}[l]{0.11\textwidth}
 \centering
      \includegraphics[height=3cm]{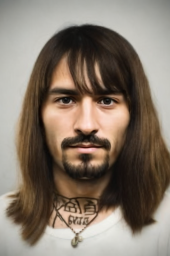}
     % \caption{Unrealistic \cite{bib_future_bias}}
 \end{subfigure}
  \begin{subfigure}[l]{0.11\textwidth}
 \centering
      \includegraphics[height=3cm]{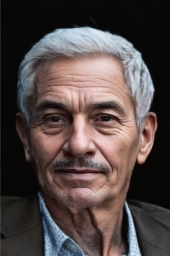}
     % \caption{Unrealistic \cite{bib_future_bias}}
 \end{subfigure}
  \begin{subfigure}[l]{0.11\textwidth}
 \centering
      \includegraphics[height=3cm]{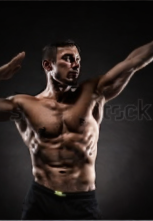}
     % \caption{Unrealistic \cite{bib_future_bias}}
 \end{subfigure}
  \caption{Biases and limitations of current state-of-the-art counterfactual generation approaches (top row): unrealistic background \cite{bib_coco_counterfactuals}, a shirtless man for a key word "tattoo" \cite{bib_social_counterfactuals}, a black and white photo for a key word "old" \cite{bib_social_counterfactuals}, unrealistic content \cite{bib_future_bias}. Counterfactuals generated using our method (bottom row) do not suffer from such biases and limitations.}
  \label{fig_main}
  \vspace{-0.4cm}
\end{figure}
Foundation models are rapidly becoming the backbone of many different types of visual analysis systems. As these models are widely available through their released weights, their adoption has spread rapidly, serving as powerful feature extractors, few-shot learners, or transfer learning bases across countless downstream tasks. However, foundation models carry biases in their encoded knowledge inherited from their training data. The datasets used for training these models are vast collections of web-scraped images that are typically undisclosed and too large and complex to be properly filtered or assessed for problematic patterns that lead to bias. When these models are used to bootstrap downstream applications\textemdash including high-stakes ones, their encoded biases silently propagate with undesirable effects during deployment. For example, one proven example of this kind of bias is reduced recognition performance for female professionals \cite{bib_sirotkin}. This presents a critical challenge: how can we detect and measure biases embedded in model weights without access to their original training distribution?

Counterfactual analysis\textemdash examining how model predictions change when protected attributes are varied\textemdash has emerged as a promising approach for detecting biases in these models \cite{bib_model_diagnosis}. It has been fostered by recent breakthroughs in generative image models. In particular, the advent of powerful text-to-image models like Stable Diffusion \cite{bib_stable_diffusion}, DALL-E \cite{bib_dalle} and Imagen \cite{bib_imagen} has revolutionized the creation of high-quality synthetic images. Whereas these models are primarily used for creative applications and generating training data in scenarios with limited ground-truth data \cite{bib_task2sim, bib_syn2, bib_syn3, bib_syn4, bib_syn5}, the quality of the generated images and theoretical control capabilities also make them valuable tools for generating counterfactual examples to study model behavior \cite{bib_social_counterfactuals, bib_coco_counterfactuals}. However, the control mechanisms required to change only protected attributes can result in reduced image diversity or introduce artifacts that affect models' predictions (see examples in Figure \ref{fig_main}). Numerous studies have shown that the generated synthetic data can contain gender, racial, and other types of social bias \cite{bib_abhishek, bib_dall_eval, bib_gender_representation, bib_biases_gen_models, bib_biases_4, bib_biases_5}. Thus, these tools risk introducing their own learned stereotypical associations into counterfactual images, such as confounding changes in clothing, pose, and environmental context (see again examples in Figure \ref{fig_main}).  When used for bias profiling, this \textit{bias leakage} makes it difficult to disentangle whether changes in the profiled model's responses stem from its own inherent biases or from stereotypical patterns introduced in the counterfactual generation.

The expression of bias in text-to-image models is often triggered by key words indicating cultural context in the text prompts \cite{bib_gen_fail, bib_future_bias}. While the fundamental long-term solution to this problem lies in better practices for dataset collection, curation and generative model design, in this paper we address this in the ``here and now'' for existing models. We present a novel approach to perform controlled attribute manipulation, through the integration of two specialized deep learning models for creating an automated human masking technique. The so-generated mask is used to constrain a SOTA local image in-painting method \cite{bib_image_inpainting}. Together, this intuitive but unexplored combination constraints image modifications to only those regions strictly necessary for changing protected attributes\textemdash such as faces and hands\textemdash while preserving all other context. This is achieved by  also imposing strict constraints on the generative model: using simple prompts that are unlikely to trigger learned biases and conditioning on the original image content.

Our approach represents a significant improvement over existing counterfactual generation methods by minimizing \textit{bias leakage} through secondary elements like clothing, pose, and environment (see Figure \ref{fig_main}). Unlike full image re-generation or manual balancing of real samples, our method significantly diminishes the probability of information leakage through secondary concepts present in the image \cite{bib_balanced_datasets}, since the contextual information for the counterfactuals is essentially identical, including the clothing and background objects. This approach allows focusing on the measurement of how protected attributes affect model predictions, and facilitates the creation of balanced datasets that maintain consistent context across protected-attribute variations. We demonstrate the effectiveness of the proposed approach through extensive experiments focused on gender as a case study, while maintaining generalizability to other protected attributes. Specifically, we apply this method to the Conceptual Captions \cite{bib_cc3m} dataset and balance its distribution of genders by in-painting the areas of persons present in the images not covered by clothing, hair, or other objects. The generated dataset is available at \nolinkurl{hidden for review}. 

We validate the potential of the so-created dataset through a three stage process. First, we evaluate the quality of our generated counterfactuals through three sets of complementary metrics: aesthetic quality and feature distribution alignment. Reported results indicate that images generated with our protocol are closer to real ones than alternative counterfactual generation approaches. Second, we validate their effectiveness for bias profiling by showing that biases identified using our counterfactuals in a known-to-be-biased model more closely match those found using real images than alternative methods, suggesting minimal introduction of additional biases. Finally, we demonstrate their utility for bias mitigation through ablated fine-tuning experiments: we first verify that models fine-tuned with our counterfactuals maintain performance on non-human-centric tasks, then show significant gender bias reduction when fine-tuning with different synthetically-balanced datasets while maintaining model performance. Specifically, our contributions are:
\begin{itemize}
\item A novel controlled counterfactual generation approach that minimizes \textit{bias leakage} by constraining modifications to essential attribute-specific regions through automated masking, yielding a more reliable bias profiling than existing approaches. 
\item A comprehensive validation framework showing that the distribution of our  counterfactuals better aligns with that of real images.
\item  Empirical evidence that our counterfactuals also enable effective bias mitigation through fine-tuning while maintaining model performance on both human-centric and general vision tasks.
\end{itemize}
\section{Related work} \label{sec:related_work}
\subsection{Image generation}
\begin{figure*}[ht!]
  \centering
\begin{subfigure}[l]{0.19\textwidth}
\centering
      \includegraphics[height=2.8cm]{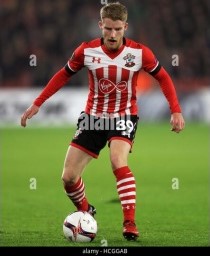}
 \end{subfigure}
 \hfill
  \begin{subfigure}[l]{0.19\textwidth}
  \centering
      \includegraphics[height=2.8cm]{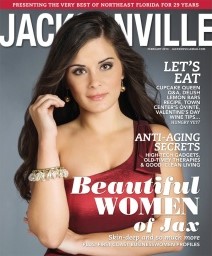}
 \end{subfigure}
   \hfill
  \begin{subfigure}[l]{0.19\textwidth}
  \centering
      \includegraphics[height=2.8cm]{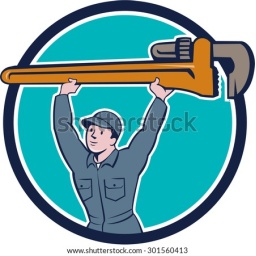}
 \end{subfigure}
   \hfill
  \begin{subfigure}[l]{0.19\textwidth}
  \centering
      \includegraphics[height=2.8cm]{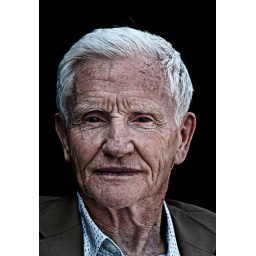}
 \end{subfigure}
    \hfill
  \begin{subfigure}[l]{0.19\textwidth}
  \centering
      \includegraphics[height=2.8cm]{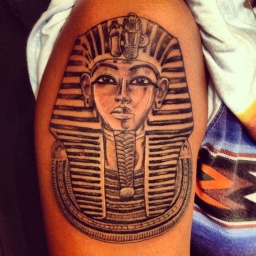}
 \end{subfigure}
 \vfill
 \begin{subfigure}[l]{0.19\textwidth}
 \centering
      \includegraphics[height=2.8cm]{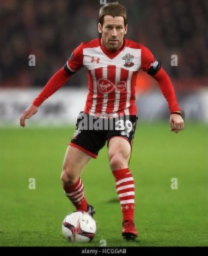}
 \end{subfigure}
 \hfill
  \begin{subfigure}[l]{0.19\textwidth}
  \centering
      \includegraphics[height=2.8cm]{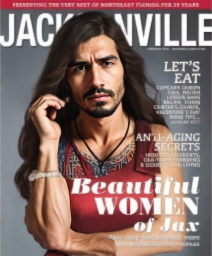}
 \end{subfigure}
   \hfill
  \begin{subfigure}[l]{0.19\textwidth}
  \centering
      \includegraphics[height=2.8cm]{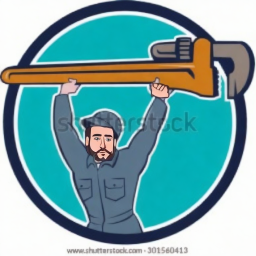}
 \end{subfigure}
   \hfill
  \begin{subfigure}[l]{0.19\textwidth}
  \centering
      \includegraphics[height=2.8cm]{figures/examples_cropped/000015371_man.png}
 \end{subfigure}
 \hfill
  \begin{subfigure}[l]{0.19\textwidth}
  \centering
      \includegraphics[height=2.8cm]{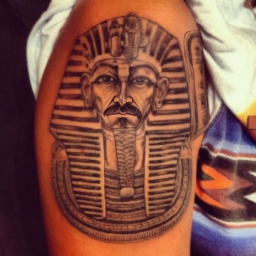}
 \end{subfigure}
 \vfill
 \begin{subfigure}[l]{0.19\textwidth}
 \centering
      \includegraphics[height=2.8cm]{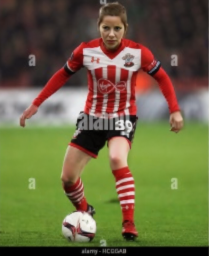}
 \end{subfigure}
 \hfill
  \begin{subfigure}[l]{0.19\textwidth}
  \centering
     \includegraphics[height=2.8cm]{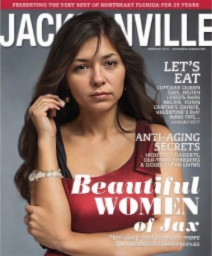}
 \end{subfigure}
   \hfill
  \begin{subfigure}[l]{0.19\textwidth}
  \centering
            \includegraphics[height=2.8cm]{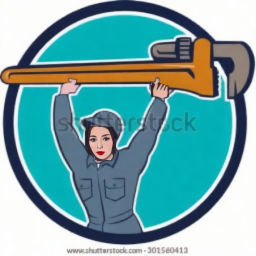}
 \end{subfigure}
   \hfill
  \begin{subfigure}[l]{0.19\textwidth}
  \centering
            \includegraphics[height=2.8cm]{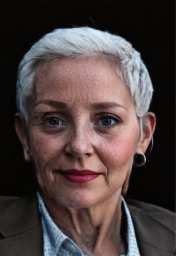}
 \end{subfigure}
 \hfill
  \begin{subfigure}[l]{0.19\textwidth}
  \centering
      \includegraphics[height=2.8cm]{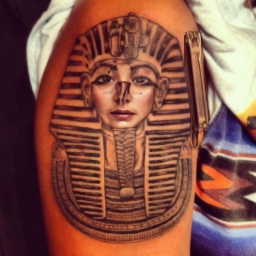}
 \end{subfigure}
 \vfill
 
  \caption{Examples of in-painted images from CC3M dataset \cite{bib_cc3m}. Top row: original image, middle row: in-painted images of men, bottom row: in-painted images of women.}
  \label{fig_examples}
  \vspace{-0.3cm}
\end{figure*}
% Synthetic data to improve performance
The use of synthetic images in computer vision is now a consolidated trend in spite of their potential limitations \cite{shumailov2024ai}. Their use has mainly focused on augmenting limited training data or generating new training samples \cite{bib_medical_synthetic, bib_noel1, bib_noel2}. Several studies have demonstrated the effectiveness of this approach across different domains. For instance, synthetic LiDAR data has proven valuable in improving semantic segmentation performance \cite{bib_montalvo}, while in infrastructure maintenance, synthetic image generation has enhanced automatic sewer pipe defect detection \cite{bib_pipe}. Regarding human-target applications, the primary challenge shifts from data scarcity to data diversity \cite{bib_dataset_diversity}. Research in this domain predominantly focuses on addressing imbalances in demographic attributes such as gender, age, and skin tone. Traditional approaches have explored methods like balanced sampling by gender or adversarial removal of gender-related information \cite{bib_balanced_datasets}. More recently, advanced generative approaches have emerged to tackle these challenges. For instance, researchers have explored prompt-to-prompt image editing for counterfactual generation \cite{bib_coco_counterfactuals, bib_orompt_to_prompt}, though this method produces fully synthetic images without specifically targeting demographic attributes.

Among image generative approaches, those strongly related to our method include a recent study proposing an in-painting method based on automatically generated captions to edit images of persons across different genders \cite{bib_gen_iccv}, and one proposing to generate synthetic counterfactuals by employing Stable Diffusion with cross-attention to guide the sampling of image batches that vary in gender and ethnicity, while maintaining context \cite{bib_social_counterfactuals}. Neither of these methods addresses how contextual elements (\textit{e.g} clothing, poses, or even background colors) might express the model's implicit biases \cite{bib_diffusion_bias_1, bib_diffusion_bias_2, bib_diffusion_bias_3}. 

In our work, we demonstrate that contextual elements should not be disregarded when generating counterfactuals, as they propagate existing or new bias. By constraining the counterfactual generation to result from localized interventions in the images we not only avoid the creation of new bias, but also generate synthetic data that can lead to superior performance after fine-tuning biased models.

\subsection{Bias estimation} \label{section_related_bias}

The detection of social biases in vision models can be broadly categorized into two main branches. The first branch focuses on direct performance measurements \cite{bib_facet, bib_overwriting_biases, bib_men_shopping, bib_equality_of_opportunity}, comparing model behavior across different values of protected attributes (e.g., male doctors vs female doctors). In this regard, researchers have analyzed various metrics including recall \cite{bib_facet}, False Positive Rate (FPR) \cite{bib_overwriting_biases}, and equalized odds \cite{bib_equality_of_opportunity} to quantify such biases. 
The second branch examines correlations in feature embeddings, operating under the hypothesis that biases propagate to downstream tasks \cite{bib_biases_ieat, bib_sirotkin}. These methods typically analyze embeddings of carefully curated image samples representing specific semantic concepts such as geographical origin or ethnicity, ages, valence, and tools \cite{bib_biases_ieat, bib_sirotkin, bib_markedness}. More sophisticated techniques in this category manipulate latent vectors along specific directions, for instance, to analyze perceptual light and hue angle variations in skin tones \cite{bib_beyond_skin_tone}.

While our approach does not relate directly to bias estimation, the proposal of a guidance mechanism to constrain the counterfactual interventions to be localized provides the basis for novel ways to mitigate the effects of known biases, as demonstrated by our results on bias profiling (see Section~\ref{sec:results}).
\section{Methodology} \label{sec:methodology}
% The following subsections explain in detail the image editing approach as well as the setups of experiments outlined in Section \ref{sec:results}.
% \subsection{Data generation framework} 

In this work, we propose an automated text-image editing framework for in-painting protected attributes of people. Overall, the data editing approach consists of mask generation and attribute inpainting, followed by caption editing to reflect the attribute changes in the text description.

\paragraph{Mask generation} 
The effectiveness of our approach emerges from a mask generation process that specifically targets attribute-relevant areas while preserving context. Previous works \cite{bib_gen_iccv} typically mask and in-paint the entire person, which inadvertently allows generative models to express their inherent biases through contextual elements. In contrast, our method employs a two-stage masking process that precisely isolates only the areas where protected attributes are physically manifested. Specifically, mask generation proceeds in three stages:

\begin{itemize}
    \item For each unlabeled input image, we first apply a specialized pre-trained ResNet-101 person segmentation model pretrained on COCO \cite{bib_coco} and Pascal VOC \cite{bib_pascal} datasets to obtain a binary person mask $\mathcal{M}_p$.
    \item  For images with non-empty $\mathcal{M}_p$, i.e., when at least one person is detected, we consecutively apply a custom ResNet-101 skin segmentation model pretrained on the images from COCO \cite{bib_coco} annotated with skin segmentation masks to generate a binary skin area mask $\mathcal{M}_s$.
    \item The final mask for an image is computed as the intersection of these two masks: $\mathcal{M} = \mathcal{M}_p \cup \mathcal{M}_s$ \footnote{we apply a morphological dilation process with a square $3 \times 3$ structured element to account for small inaccuracies in the segmentation process}.
\end{itemize}

We have also explored the use of the Segment Anything Model (SAM) \cite{bib_sam}. However, empirically it produced worse results than a cascade of smaller specialized models (see  qualitative examples in the supplementary material).

\paragraph{In-painting}
For the attribute modification step, we employ BrushNet \cite{bib_brushnet}, a state-of-the-art dual-branch in-painting model that enables localized image modifications. BrushNet separates the processing of masked image features from noisy latent variables, permitting fine-grained control over the in-painting process. The model is fed by an input image, an associated binary mask $\mathcal{M}$, and a text prompt in the format: \textit{``A photo of a \{value of a protected attribute\}''}. The generation process is conditioned by all three. The choice of BrushNet as our core in-painting module is based on its reported superior quantitative and qualitative performance when compared to such models as Blended Latent Diffusion \cite{bib_bld}, ControlNet \cite{bib_control_net}, HD-Painter \cite{bib_hd_painter} and PowerPaint \cite{bib_power_paint}.

\paragraph{Captions editing}

To update captions to reflect new values of protected attributes\textemdash such as gender, we identify and replace words that explicitly or implicitly indicate the attribute’s original value. For example, terms like \textit{woman}, \textit{girl}, \textit{mother}, \textit{man}, or \textit{boy} explicitly state gender, while others like \textit{queen}, \textit{waitress}, or \textit{businessman} imply it. We establish corresponding counterparts for these terms and substitute them with those representing the opposite gender. For example, \textit{Man$\xrightarrow{}$Woman}, \textit{Actress$\xrightarrow{}$Actor}, \textit{King$\xrightarrow{}$Queen}. If a term is gender-neutral (e.g., \textit{person}, \textit{human}, \textit{artist}), no changes are made. Refer to the supplementary material for a list of correspondences.
\section{Verification of the proposed approach} \label{section_verification}

\begin{table}[]
\centering
\addtolength{\tabcolsep}{-0.3em}

\begin{tabular}{@{}cccccc@{}}
\toprule
           Metric & Real  & Ours & SD \cite{bib_future_bias} & SC \cite{bib_social_counterfactuals} & CC \cite{bib_coco_counterfactuals} \\ \midrule
           HPS \cite{bib_hps} $\uparrow$              & 0.12& 0.11&{\color[HTML]{036400} 0.17}&    0.16& 0.11       \\
           AS \cite{bib_ar} $\uparrow$                                                     & 3.76& 3.61& 3.79&    {\color[HTML]{036400} 4.49}& 3.69       \\
           IR \cite{bib_ir} $\uparrow$                     & -2.12& -2.20& {\color[HTML]{036400} -1.41}&    -2.11&  -2.19      
  %CLIP \cite{bib_clip_sim} $\uparrow$ &
  %&
  %&
  %&
  % &
   \\
           FID \cite{bib_fid} $\downarrow$                & -     & {\color[HTML]{036400} 17.02}& 32.59 & 54.86   &  39.28      \\
           KID \cite{bib_demystifying_gans} $\downarrow$ & -     & {\color[HTML]{036400} 0.008} & 0.017 &  0.045  &  0.022      \\

  CMMD \cite{bib_ccmd} $\downarrow$ &
  - &
  0.359 & 0.25
  & 0.496
   & 0.411
   \\
\end{tabular}
\caption{Quantitative evaluation of the realism of real images, inpainted images (ours), fully synthetic images - \textbf{SD} (Stable Diffusion 1.5, using captions from CC3M \cite{bib_cc3m}) \cite{bib_future_bias}, Social Counterfactuals \cite{bib_social_counterfactuals} - \textbf{SC} (fully synthetic batches expressing different values of protected attributes) and COCO Counterfactuals \cite{bib_coco_counterfactuals} - \textbf{CC} synthetic counterfactuals based on the edited COCO-MS \cite{bib_coco} captions. Note that since FID, KID and CCMD compare the similarities of distributions of feature embeddings, we compute them for the pairs [Real; Synthetic].}
\label{table_image_realism}
\vspace{-0.4cm}
\end{table}

\subsection{Experimental Setup}

To validate the proposed framework, we apply it to a large text-image dataset ``Conceptual Captions'' (CC3M) \cite{bib_cc3m} by inpainting genders (men and women onto the images containing people). At the time of writing this paper, only 2.13M images were accessible through the links provided in CC3M. Among those, 606,041 images were identified as containing people and used to create synthetic counterfactual samples, totaling 1,212,082 images. During image generation, to promote diversity, each image in-painting process employs a unique random seed, while all images are processed at the model's native $768\times768$ pixel resolution.

We evaluate the changes in downstream performance when a CLIP model is pre-trained fully or partially on synthetic data. To this end, we gradually replace real images containing people with couples of their synthetic versions with both genders in-painted (see Figure \ref{fig_examples}). Following this approach, we obtain 3 versions of the CC3M dataset: original CC3M, CC3M with half of the images of people replaced and CC3M with all images of people replaced\footnote{Note, that `all' refers to all images in which people have been detected by the segmentation models.}. We denote the three versions as CC3M ($0\%$), CC3M ($50\%$) and CC3M ($100\%$), respectively\footnote{Here, $0\%$, $50\%$, and $100\%$ of images of people account for $0\%$, $14.24\%$, and $28.49\%$ of total images in CC3M, respectively.}.
We  consider two backbone architectures for CLIP models: ViT-B/16 and ResNet-50. The models are trained from scratch and the last checkpoint is then evaluated in the ImageNet zero-shot classification setting. This training regime replicates the previously reported setups \cite{bib_future_bias}.

In all experiments presented in Sections \ref{sec_model_profiling} and \ref{sec_downstream}, we use a ViT-B/16 CLIP \cite{bib_openclip} model that was pretrained on LAION-2B~\cite{bib_laion5b}~\footnote{Model weights are available in the \href{https://github.com/mlfoundations/open_clip}{OpenCLIP} repository}. Fine-tuning (see Section \ref{sec_downstream}) is done for 30 epochs with warmup and a maximum learning rate of 5e-5, following the recently proposed recommendations for fine-tuning CLIP models \cite{bib_finetuning_clip}.

% [To do: describe the models that have been used, how is the generation process carried out (any parameters?), how the fine-tuning is performed (parameters, learning rates, augmentations, protocols,...]

\subsection{Quality Assessment of the Counterfactuals}  \label{sec_aesthetic_metrics}

Over the years, many metrics have been proposed for assessments of the quality of generated images (i.e. their similarity to real images). Generally, they rank each image's quality using a model trained on human assessments of synthetic images (aesthetic quality evaluation) or the distance between the distributions of generated and ground truth (reference) images (feature distribution analysis).
% We include the results for the real CC3M images \cite{bib_cc3m}, fully synthetic images (not in-painted) generated from the original CC3M captions through Stable Diffusion, and other state-of-the-art counterfactual generation approaches \cite{bib_social_counterfactuals, bib_coco_counterfactuals} (see Section \ref{sec_realsim_results}).
\vspace{-0.4cm}
% (Image Reward \cite{bib_ir}, HPS-v2 \cite{bib_hps}, Aesthetic Score \cite{bib_ar}) or compute the distance between the distributions of generated and ground truth (reference) images (FID \cite{bib_fid}, KID \cite{bib_demystifying_gans}, CCMD \cite{bib_ccmd}). Alternatively, the difference between the generated captions  for real (reference) images and the synthetic counterfactuals can be compared with a metric like BERT Score \cite{bib_bertscore}. Since our counterfactuals are in-painted and a part of the image is kept unchanged, for all experiments on aesthetic quality evaluation and distribution analysis, we only analyze the parts of images modified by a generative model (to avoid the evaluation metrics favoring our approach due to the presence of real data in the modified images), i.e. the areas of people not covered by clothing. To ensure a fair comparison, we follow the same protocol for other counterfactual generation methods. Finally, the BERT Scores are computed for full images because the image captions usually describe the context and the person in the image.

\paragraph{Aesthetic Quality Evaluation.}
We use a range of image realism metrics, including the most recent state-of-the-art approaches (Image Reward \cite{bib_ir}, HPS-v2 \cite{bib_hps}, Aesthetic Score \cite{bib_ar}) to verify the realism of generated images. We include the results for the real CC3M images \cite{bib_cc3m}, fully synthetic images (not in-painted) generated with Stable Diffusion \cite{bib_future_bias}, and the state-of-the-art counterfactual generation approaches \cite{bib_social_counterfactuals, bib_coco_counterfactuals}.
\vspace{-0.4cm}
\paragraph{Feature Distribution Analysis.}
Additionally, we compute the distances between the features extracted from the real and synthetically generated data, using FID \cite{bib_fid}, KID \cite{bib_demystifying_gans}, and recently proposed CMMD \cite{bib_ccmd}.

% \paragraph{Caption-based Evaluation}
% Finally, we assess the similarity of captions generated for inpainted and reference images. To this aim, we apply a captioning model (ClipCap \cite{bib_clipcap}) to the real and in-painted images (following an established protocol \cite{bib_future_bias}), and compute their similarity using  BERT Score \cite{bib_bertscore}.
% This evaluation approach corresponds to direct assessments of the quality of generated images (i.e. their similarity to real images), and could use traditional metrics such as FID \cite{bib_fid}, KID \cite{bib_demystifying_gans}, or newer state-of-the-art methods such as Image Reward \cite{bib_ir} and others. In this work, we choose to focus on the metrics that allow the evaluation of image quality on a per-instance basis, limiting the choice to Image Reward \cite{bib_ir}. HPS-v2 \cite{bib_hps}, Aesthetic Score \cite{bib_ar} and CLIP similarity \cite{bib_clip_sim}.

\subsection{Model profiling} \label{sec_model_profiling}

\paragraph{Bias identification measures.}
Since we operate with CLIP models that easily allow zero-shot inference, we only focus on the bias detection methods that directly measure models' performance (see the discussion in Section \ref{section_related_bias}):

\begin{itemize}
    \item \textbf{Person Preference} \cite{bib_markedness} measures how often a CLIP model ``prefers'' the caption \textit{``A photo of a person''} to a caption \textit{``A photo of a \{value of a protected attribute\}''}. For example, person preference returns the maximum  score of $1.0$ if for all images of women the model has a higher cosine similarity between the embedding of each image and a prompt \textit{``A photo of a person''} rather than a prompt \textit{``A photo of a woman''}. We compute person preference scores on the MORPH dataset \cite{bib_morph}.
    \item \textbf{Self-Similarity Score (Markedness)} \cite{bib_markedness} computes the mean cosine similarity between embeddings of image samples representing a group (e.g., individuals with a light skin-tone, young age, blonde hair). In a binary case, higher values of self-similarity for one group indicate its lower feature diversity, implying existence of a bias. In the ideal case, self-similarity of both groups is equal. We compute self-similarity scores on the FairFace dataset \cite{bib_morph}.
    \item \textbf{Equality of Opportunity} \cite{bib_equality_of_opportunity} measures the true positive rate (recall) on a task for groups of people with different values of a protected attribute. We compute equality of opportunity scores on the occupations from the FACET dataset \cite{bib_facet} that are present in CC3M (refer to the supplementary material for the methodology and the list of studied occupations).
    \item \textbf{Gender classification} reports recall for each gender.
    \end{itemize}

In all our experiments, no downstream task training is performed and inference is performed in the zero shot mode to prevent the imbalances and implicit biases present in downstream datasets \cite{bib_dataset_biases_1, bib_dataset_biases_2} from obfuscating the results of the bias metrics.

\paragraph{Analysis of a biased model.}
We evaluate the possibility of using synthetically generated counterfactuals to detect biases in trained models. To this end, we compare the results of the bias identification measures computed for the real and synthetic data (see Table \ref{table_profiling}). If the differences between them are small, it can be said that the synthetic counterfactuals can serve as a viable tool for bias profiling.

\subsection{Bias Mitigation through Fine-tuning} \label{sec_downstream} 
\begin{table}[]
\centering
\addtolength{\tabcolsep}{-0.1em}
\begin{tabular}{@{}ccccccccccc@{}}
\toprule
                                                     & c1 & c2 & c3 & c4     & c5   & c6     & c7   & c8 & c9     & c10  \\ \midrule
RW &
  \checkmark &
   &
  \checkmark &
  \checkmark &
  \checkmark &
   &
   &
  \checkmark &
   &
   \\
RM &
   &
  \checkmark &
  \checkmark &
   &
   &
  \checkmark &
  \checkmark &
  \checkmark &
   &
   \\
SW &    &    &    &        &      & C & S & *  & C & S \\
SM   &    &    &    & C & S &        &      & *  & C & S \\ \bottomrule
\end{tabular}
\caption{Reference codes of data partitions used in the experiments presented in Section \ref{sec_results_finetuning}. In each row, \textbf{R} denotes \textbf{R}eal, \textbf{S}\textemdash \textbf{S}ynthetic, \textbf{M}\textemdash \textbf{M}en, \textbf{W}\textemdash \textbf{W}omen, \textbf{C} means a gender \textbf{c}hange w.r.t the source image, \textbf{S} means that a gender is kept the \textbf{s}ame. For instance, column ``c4'' describes real images of women and synthetic images of men generated using the real images of women as a reference. In ``c8'', * refers to inclusion of both \textbf{C} and \textbf{S}.}
\label{table_splits}
\vspace{-0.5cm}
\end{table}

We evaluate the effects of using synthetic data on non human-centric tasks, by measuring the zero-shot accuracy on ImageNet-1k \cite{bib_imagenet} for models pretrained on real and synthetic images. Since ImageNet-1k does not explicitly contain classes related to people (although people do appear in the dataset), we expect to not see a drop in performance.

Finally, we evaluate the effects on fine-tuning the LAION-2B pretrained CLIP model with our in-painted images. To this end, we design a series of experiments in which we fine-tune the model on various combinations of real and in-painted images of people of two genders. Specifically, we use the subsets of CC3M presented in Table \ref{table_splits}.

\section{Results} \label{sec:results}

Overall, 1.2+M image-caption pairs have been generated applying the framework presented in Section \ref{sec:methodology}. Some examples are shown in Figure \ref{fig_examples} and in Supplementary material. The following sections provide a summary and an analysis of the experimental results obtained with the synthetic data.

\subsection{Does pinpoint counterfactual generation reduce image realism?} \label{sec_realsim_results}

\begin{table}[]
\centering
\begin{tabular}{@{}ccc}
\toprule
Dataset &
  Self-similarity $\downarrow$ &
  Gender classification $\uparrow$\\ \midrule
Real - M & 0.433 & 78.39 \\
Real - W & 0.504 & 92.82 \\
 $|\Delta_{MW}^{real}|$& 0.071& 14.43\\ \midrule
Ours - M & 0.525 & 92.25 \\
Ours - W & 0.592 & 97.22 \\
 $|\Delta_{MW}^{ours}|$& 0.067& 4.97\\ \midrule
SC - M   & 0.653 & 99.46 \\
SC - W &
  0.672 &
  99.92 \\
 $|\Delta_{MW}^{SC}|$& 0.019& 0.46\\
\end{tabular}
\caption{Bias profiling using real images (first two rows -  CC3M \cite{bib_cc3m}), and images obtained with our propsed approch (middle two rows), and a state-of-the-art counterfactual generation method \cite{bib_social_counterfactuals} (bottom two rows) on a LAION-pretrained  \cite{bib_laion5b} ViT-B/16 CLIP model. $|\Delta_{MW}^{dataset}|$ show the absolute value of a difference between the two gender groups for each metric.}
\label{table_profiling}
\vspace{-0.5cm}
\end{table}

% \begin{table}[]
% \centering
% \begin{tabular}{@{}cccc}
% \toprule
% Dataset &
%   Self-sim. $\downarrow$ &
%   \begin{tabular}[c]{@{}c@{}}Person \\ pref. $\uparrow$\end{tabular} &
%   \begin{tabular}[c]{@{}c@{}}Gender \\ Class. $\uparrow$\end{tabular} \\ \midrule
% Real - M & 0.433 & 0.80 & 78.39 \\
% Real - W & 0.504 & 0.55 & 92.82 \\
%  $|\Delta_{MW}^{real}|$& 0.071& 0.25& 14.43\\ \midrule
% Ours - M & 0.525 & 0.70 & 92.25 \\
% Ours - W & 0.592 & 0.52 & 97.22 \\
%  $|\Delta_{MW}^{ours}|$& 0.067& 0.18& 4.97\\ \midrule
% SC - M   & 0.653 & 0.81 & 99.46 \\
% SC - W &
%   0.672 &
%   0.37 &
%   99.92 \\
%  $|\Delta_{MW}^{SC}|$& 0.019& 0.44& 0.46\\
% \end{tabular}
% \caption{Bias profiling using real images (first two rows -  CC3M \cite{bib_cc3m}), and images obtained with our propsed approch (middle two rows), and a state-of-the-art counterfactual generation method \cite{bib_social_counterfactuals} (bottom two rows) on a LAION-pretrained  \cite{bib_laion5b} ViT-B/16 CLIP model. $|\Delta_{MW}^{dataset}|$ show the absolute value of a difference between the two gender groups for each metric.}
% \label{table_profiling}
% \end{table}
We study the image realism metrics outlined in Section \ref{sec_aesthetic_metrics} of both synthetic counterfactual and real images. Table \ref{table_image_realism} summarizes the obtained results and shows a slight decrease in image realism between real images and images in-painted with our approach (according to HPS, AS and IR) or generated with prompt-to-prompt image editing with cross attention control \cite{bib_coco_counterfactuals}. However, for fully synthetic data generated with Stable Diffusion (SD) \cite{bib_future_bias} or social counterfactuals (SC) \cite{bib_social_counterfactuals}, the same metrics \textbf{indicate a level of realism that is even higher than that of the reference real images} (higher values of HPS and Image Reward for SD and Aesthetic score for SC). On the other hand, FID and KID that measure the distance between the real and generated samples score lower (implying higher realism) for our counterfactuals than all alternative approaches, whereas CMMD also ranks them high (second best among the studied ones). In the light of these numbers, we conclude that the proposed pinpoint counterfactual method does not significantly reduce image realism.

\subsection{Do pinpoint counterfactuals preserve discriminative power for general recognition?}

\begin{figure}
    \centering
    \includegraphics[width=0.5\textwidth]{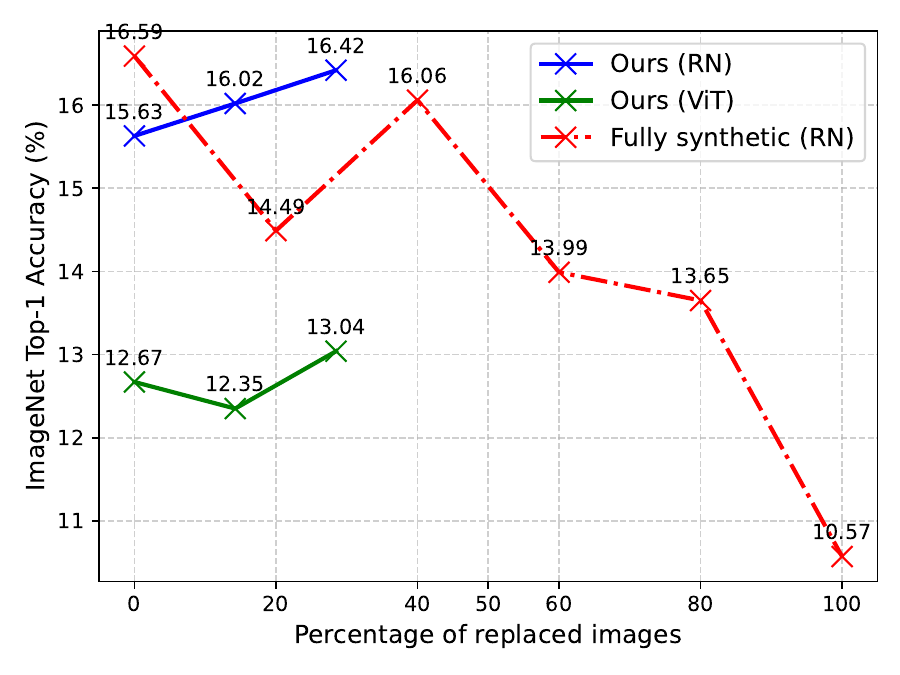}
    \caption{ImageNet zero-shot classification accuracy for ResNet-50 and ViT-B/16 models pretrained with CC3M modified using our approach and fully synthetic CC3M.}
    \label{fig_in1k}
    \vspace{-0.3cm}
\end{figure}
\begin{figure*}[t]
    \centering
    \includegraphics[width=0.8\textwidth]{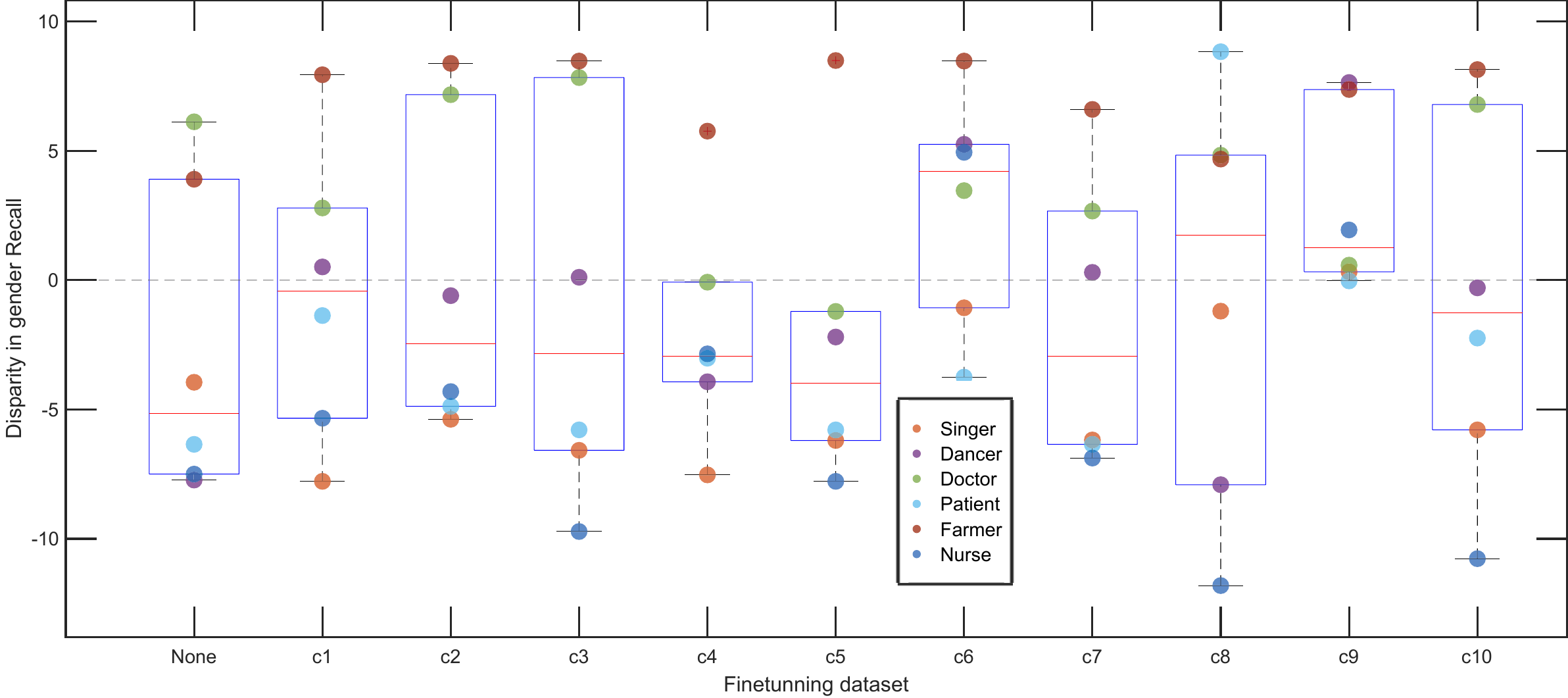}
    \vspace{-0.2cm}
    \caption{Differences in per-gender recall on select occupations of FACET dataset \cite{bib_facet}. The inference is done in zero-shot mode with ViT-B/16 CLIP pretrained on LAION-2B \cite{bib_laion5b} and fine-tuned using the combinations of real and synthetic data describe in Table \ref{table_splits}. Larger boxes indicate a larger total gender bias. Refer to supplementary material for complete data on recall.}
    \label{fig_facet}
    \vspace{-0.4cm}
\end{figure*}

To explore whether the proposed counterfactual generation method (1) avoids introducing image artifacts that could interfere with the discriminative power of models trained for general recognition non-human-centered tasks and (2) maintains diversity comparable to real images in non-human objects, we evaluate performance using the counterfactuals for ImageNet-1k zero-shot classification. Figure \ref{fig_in1k} compares ImageNet-1k zero-shot classification accuracy using pretrained models: ResNet-50 and ViT-B/16 CLIP with our pinpoint counterfactuals versus ResNet-50 with fully synthetic CC3M images \cite{bib_future_bias}. As explained in Section \ref{sec_downstream}, we use different versions of CC3M in which we gradually replace the original images containing people with the generated counterfactuals [$0\%; 50\%; 100\%$] \textemdash for our counterfactuals, or gradually replace all images (not only the ones containing people) \textemdash for the full synthetic, \cite{bib_future_bias} [$0\%; 20\%; 40\%; 60\%; 80\%; 100\%$].
As shown in Figure \ref{fig_in1k},
there is a clear downward trend in classification accuracy as more fully synthetic images are added (red line)\footnote{Note that the blue and red lines don't start at the same point because our version of CC3M does not contain all of the original images since many links are now broken.}. In contrast, gradually replacing images of people modified with our approach results in improved classification accuracy for both ResNet and ViT models\footnote{ViT-B/16 underperforms compared to ResNet-50, likely due to the size of the training dataset ($<3$M text-image pairs).}.

\subsection{Can pinpoint generated counterfactuals be used for model profiling?}

We profile a ViT-B/16 CLIP model trained with LAION-2B with real images to establish a baseline for bias profiling (see Section \ref{sec_model_profiling}). We then compute the same bias measures using our generated counterfactuals and a reference state-of-the-art alternative \cite{bib_social_counterfactuals}. Table \ref{table_profiling} summarizes the results and shows that profiling on the images generated with our method more closely replicates the results on real data, while the other approach tends to significantly underestimate the model's bias (leading to substantially smaller differences in self-similarity and gender classification between the two genders). The alignment between bias measurements on our synthetic images and real data shows promise for using pinpoint counterfactuals in model profiling, particularly in scenarios with limited or inaccessible real-world data. We hypothesize that this fidelity stems from our method's capacity to isolate attribute modifications while preventing the leakage of biases through secondary visual characteristics (see Figure \ref{fig_main}).

\subsection{Does balancing with synthetic data reduce gender bias?} \label{sec_results_finetuning}

We follow the protocol described in Section \ref{sec_downstream} to fine-tune the models with our synthetic data and measure the changes in gender bias. Results are compiled in Table \ref{table_finetuning} \textemdash which includes ImageNet-1k zero-shot perfomance for model reference, and Figure \ref{fig_facet}\textemdash that explores equality of opportunity.
\vspace{-0.8cm}
\paragraph{Person preference.} As can be seen from Table \ref{table_finetuning}, fine-tuning on real images, consistently results in imbalanced (c1, c3) or low, $<0.40$, person preference scores (c2), indicating favoring men (c1, c3). On the other hand, fine-tuning with combinations of real and our counterfactuals(c4, c6, c7, c9) leads to much more balanced and higher person preference scores, indicating reduction in model's bias.
\vspace{-0.4cm}
\paragraph{Self-similarity.} We observe that fine-tuning with our pinpoint counterfactuals or  with their combination with real data results in a more balanced self-similarity between the gender groups (from the range of $[0.02; 0.04]$ for real images down to $[0.00; 0.03]$ when using our counterfactuals), with a single exception (c10) that only contains synthetic data. Notably, in two cases (c4, c8) we are able to achieve perfectly balanced self-similarity scores that indicate a strong reduction in bias.
\vspace{-0.35cm}
\paragraph{Gender classification.} As could be expected, finetuning only on real images of men or women (c1, c2) increases the gender classification accuracy for one gender and decreases it for the opposite gender. However, at least in one case (c6), we observe that the use of our method for generating the counterfactual gender, results in balanced gender classification recalls ($96.91\%$ and $96.84\%$). 

% %conclusion of the subsection
% Hence, it can be said that in certain conditions, the use of gender-inpainting can decrease gender bias and improve the model's performance).
\vspace{-0.45cm}
\paragraph{Equality of opportunity.} Finally, we examine how different fine-tuning setups affect the model gender-occupation biases. To this end, we compute recall on the selected occupations of FACET \cite{bib_facet} (see Section \ref{sec_model_profiling}), measuring the equality of opportunity for the two genders. As can be seen from Figure \ref{fig_facet} (refer to the supplementary material for the complete results and individual recalls), the baseline model exhibits expected societal gender biases, with over $7\%$ disparity rates between the genders for some occupations (nurse, dancer). Using real balanced gender data (c3) eliminates bias for only one occupation (dancer), while having a total sum of disparity rates per occupations larger than the baseline model. On the other hand, a purely synthetic configuration (c9) almost completely eliminates gender biases for the ``doctor'', ``patient'' and ``nurse'' occupations, despite lacking real images, suggesting effective preservation of occupation-relevant features during counterfactual generation. Finally, mixed real-synthetic configurations (c4-c7) demonstrate asymmetric improvements, but result in a smaller total bias across all occupations (w.r.t the original model). This suggests counterfactual examples can expand the model's feature space while preserving representations.
    
Interestingly, configuration c8, combining all real and synthetic data, while leading to the most substantial improvements in performance for some occupations (dancer, nurse - see Table 1 in supplementary material), does not lead to the reduction in recall disparity between the genders.

\begin{table*}[ht!]
\centering
\begin{tabular}{@{}ccccccccc@{}}
\toprule
\multirow{2}{*}{Data type} &
  \multirow{2}{*}{\begin{tabular}[c]{@{}c@{}}Finetuning\\ Dataset\end{tabular}} &
  \multirow{2}{*}{\begin{tabular}[c]{@{}c@{}}IN-1K\\ zero-shot\end{tabular}} &
  \multicolumn{2}{c}{\begin{tabular}[c]{@{}c@{}}Self-similarity\\ (FairFace)\end{tabular}} &
  \multicolumn{2}{c}{\begin{tabular}[c]{@{}c@{}}Person preference\\ (MORPH)\end{tabular}}  &
  \multicolumn{2}{c}{\begin{tabular}[c]{@{}c@{}}Gender \\ classification\end{tabular}} \\ \cmidrule(l){4-9} 
 &
   &
   &
  Man &
  Woman &
  Man &
  Woman &
  Man &
  Woman \\ \midrule
\multicolumn{1}{c|}{\multirow{4}{*}{Real}} &
  None &
  70.20 &
  0.45 &
  0.48 &
  0.73 &
  0.05 &
  98.59 &
  98.42 \\
\multicolumn{1}{c|}{} &
  c1&
  69.98 &
  0.48 &
  0.52 &
  0.81 &
  0.03 &
  96.06 &
  100.00 \\
 & c2& 70.12 & 0.49 & 0.51 & 0.44 & 0.37 & 99.64 &92.11 \\
\multicolumn{1}{c|}{} &
  c3&
  69.98 &
  0.49 &
  0.53 &
  0.39 &
  0.09 &
  98.91 &
  97.37 \\ \midrule
\multicolumn{1}{c|}{\multirow{5}{*}{\begin{tabular}[c]{@{}c@{}}Real +\\ ours\end{tabular}}} &
  c4&
  69.14 &
  0.53 &
  0.53 &
  1.0 &
  0.85 &
  81.13 &
  100.00 \\
\multicolumn{1}{c|}{} &
  c5&
  69.61 &
  0.53 &
  0.55 &
  0.99 &
  0.22 &
  88.09 &
  100.00 \\
\multicolumn{1}{c|}{} &
  c6&
  69.18 &
  0.51 &
  0.54 &
  0.99 &
  0.98 &
  96.91 &
  96.84 \\
\multicolumn{1}{c|}{} &
  c7& 69.51
   & 0.53
   & 0.55
   & 0.84
   & 0.96
   & 92.37
   & 99.62
   \\
\multicolumn{1}{c|}{} &
  c8&
  63.27 &
  0.39 &
  0.39 &
  0.03 &
  0.07 &
  97.12 &
  91.18 \\ \midrule
\multicolumn{1}{c|}{\multirow{2}{*}{Ours}} &
  c9&
  67.69 &
  0.58 &
  0.59 &
  0.98 &
  0.73 &
  29.52 &
  98.29 \\
  \multicolumn{1}{c|}{} &
  c10& 68.47
   & 0.59
   & 0.55
   & 0.98
  & 0.19
   & 95.66
   & 99.08
   \\ \bottomrule
\end{tabular}
\caption{Gender bias in ViT-B/16 CLIP model pretrained on ViT-B/16 CLIP model pretrained on LAION-2B \cite{bib_laion5b} and finetuned on real or real+synthetic data generated using the proposed framework (see Section \ref{sec_results_finetuning}). Here, rM stands for the real images of men, rW - real images of women, sM - synthetic (inpainted) images of men and sW - synthetic (inpainted) images of women. See Table \ref{table_splits} for more details.}
\label{table_finetuning}
\end{table*}

% \usepackage{multirow}

% \begin{figure*}
%     \centering
%     \includegraphics[width=1.0\textwidth]{figures/Self-sim.pdf}
%     \caption{Self-similarity \cite{bib_markedness} of embeddings of images of people from the FairFace dataset \cite{bib_fairface}, extracted with a ViT-B/16 CLIP model pretrained on LAION-2B \cite{bib_laion5b} and finetuned on real or real+synthetic data generated using the proposed framework (see Section \ref{sec_results_finetuning}). \textcolor{red}{UPDATE THE FIGURE IF WE DECIDE TO KEEP IT}}
%     \label{fig_self_sim}
% \end{figure*}
% \input{sec/7_discussion}
\section{Discussion and limitations} \label{sec:limitations}
In this work, we focus on creating a framework for generating balanced datasets to enable comprehensive analysis of social bias in foundation vision and vision-language models. Our approach preserves real distributions of concepts not explicitly expressing protected attributes while adding diversity to the protected attribute distributions themselves. This selective modification is evidenced by the maintained performance on generic recognition tasks such as ImageNet zero-shot classification (see Figure \ref{fig_in1k}).

Experiments reveal several interesting findings about current vision models. First, many state-of-the-art image realism metrics fail to correctly assess synthetic image realism, often scoring real data lower than synthetic alternatives (see Table \ref{table_image_realism}). We hypothesize this stems from these metrics over-optimizing for local image features while failing to capture higher-level semantic information. Additionally, our finetuning experiments in Section \ref{sec_results_finetuning} suggest that CLIP models may learn to associate gender primarily with contextual cues and clothing rather than physiological characteristics. This is particularly evident in setups c4 and c5 (Table \ref{table_finetuning}), where gender switching leads to performance degradation across ImageNet-1k and gender classification.

\begin{figure}[ht!]
  \centering
  
 \begin{subfigure}[l]{0.10\textwidth}
    \centering
      \includegraphics[height=2cm]{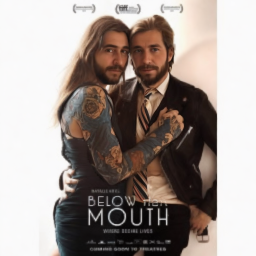}
     \caption{Same gender.}
     \label{fig_limitations_same_gender}
 \end{subfigure}
\begin{subfigure}[l]{0.15\textwidth}
\centering
      \includegraphics[height=1.8cm]{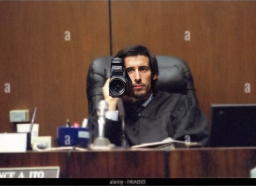}
     \caption{Camera.}
     \label{fig_limitations_camera}
 \end{subfigure}
 \begin{subfigure}[l]{0.15\textwidth}
 \centering
      \includegraphics[height=1.8cm]{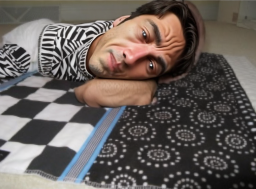}
     \caption{Horizontal pose.}
     \label{fig_limitations_pose}
 \end{subfigure}
  \caption{Limitation of the proposed approach: inpainting of the same gender for all people in the image (a), inpainting of a camera (b), distorted face for a person in a non-vertical pose (c). See Section \ref{sec:limitations} for a discussion on the limitations.}
  \label{fig_limitations}
  \vspace{-0.3cm}
\end{figure}

In spite of these advantages, our method faces several limitations. While we control for stereotypical contextual attributes through BrushNet-guided inpainting \cite{bib_gen_fail}, subtle generative model biases may persist in the inpainted areas, such as makeup for women or beards for men, reflecting the attribute distribution of the generative model's pretraining dataset. We also observe that when multiple people are present in an image, they are all edited to have the same gender, reducing intra-image gender variance (see Figure \ref{fig_limitations_same_gender}). Finally, while rare, certain generative artifacts appear in specific scenarios, such as distorted faces in non-vertical poses (Figure \ref{fig_limitations_pose}) and inexplicable additions of digital cameras (Figure \ref{fig_limitations_camera}), which we hypothesize may stem from detailed prompts in the pretraining data.

\section{Conclusion}

% In this work, we present a novel in-painting framework for counterfactual image generation that minimizes bias leakage by preserving the original contextual information. We demonstrate the effectiveness of our approach by creating a gender-balanced version of a large dataset, CC3M, and using the generated data for both identifying and mitigating gender bias in pretrained state-of-the-art models. While we focus on gender biases in this study, our approach can be adapted for any other protected attribute, including ethnicity, age, skin-tone, etc. Finally, we demonstrate that using the images edited with the proposed approach does not lead to the performance degradation on non human-centric tasks.

In this work, we have presented a counterfactual generation framework that constrains attribute modifications to localized regions while preserving contextual information. Extensive experiments demonstrate that our approach generates realistic counterfactuals while avoiding the introduction of new biases through contextual elements. The proposed method enables accurate bias profiling through minimal interference with non-attribute features, and potential for bias mitigation through fine-tuning, as evidenced by balanced performance across protected groups with preserved general vision capabilities. Our strategy is extendable to other protected attributes, offering a systematic framework for analyzing and addressing bias in foundation models. This work provides a foundation for future research in controlled counterfactual generation and bias mitigation.

{
    \small
    \bibliographystyle{ieeenat_fullname}
    \bibliography{main}

\begin{thebibliography}{66}
\providecommand{\natexlab}[1]{#1}
\providecommand{\url}[1]{\texttt{#1}}
\expandafter\ifx\csname urlstyle\endcsname\relax
  \providecommand{\doi}[1]{doi: #1}\else
  \providecommand{\doi}{doi: \begingroup \urlstyle{rm}\Url}\fi

\bibitem[Anderson et~al.(2021)Anderson, Shrivastava, Truong, Majumdar, Parikh, Batra, and Lee]{bib_syn2}
Peter Anderson, Ayush Shrivastava, Joanne Truong, Arjun Majumdar, Devi Parikh, Dhruv Batra, and Stefan Lee.
\newblock Sim-to-real transfer for vision-and-language navigation.
\newblock In \emph{Conference on Robot Learning}, pages 671--681. PMLR, 2021.

\bibitem[Avrahami et~al.(2023)Avrahami, Fried, and Lischinski]{bib_bld}
Omri Avrahami, Ohad Fried, and Dani Lischinski.
\newblock Blended latent diffusion.
\newblock \emph{ACM transactions on graphics (TOG)}, 42\penalty0 (4):\penalty0 1--11, 2023.

\bibitem[Bansal et~al.(2022)Bansal, Yin, Monajatipoor, and Chang]{bib_biases_5}
Hritik Bansal, Da Yin, Masoud Monajatipoor, and Kai-Wei Chang.
\newblock How well can text-to-image generative models understand ethical natural language interventions?
\newblock In \emph{Proceedings of the 2022 Conference on Empirical Methods in Natural Language Processing}, pages 1358--1370, 2022.

\bibitem[Bianchi et~al.(2023)Bianchi, Kalluri, Durmus, Ladhak, Cheng, Nozza, Hashimoto, Jurafsky, Zou, and Caliskan]{bib_biases_gen_models}
Federico Bianchi, Pratyusha Kalluri, Esin Durmus, Faisal Ladhak, Myra Cheng, Debora Nozza, Tatsunori Hashimoto, Dan Jurafsky, James Zou, and Aylin Caliskan.
\newblock Easily accessible text-to-image generation amplifies demographic stereotypes at large scale.
\newblock In \emph{Proceedings of the 2023 ACM Conference on Fairness, Accountability, and Transparency}, pages 1493--1504, 2023.

\bibitem[Bi{\'n}kowski et~al.(2018)Bi{\'n}kowski, Sutherland, Arbel, and Gretton]{bib_demystifying_gans}
M Bi{\'n}kowski, DJ Sutherland, M Arbel, and A Gretton.
\newblock Demystifying mmd gans.
\newblock ICLR, 2018.

\bibitem[Brielmann et~al.(2022)Brielmann, Buras, Salingaros, and Taylor]{bib_dataset_diversity}
Aenne~A Brielmann, Nir~H Buras, Nikos~A Salingaros, and Richard~P Taylor.
\newblock What happens in your brain when you walk down the street? implications of architectural proportions, biophilia, and fractal geometry for urban science.
\newblock \emph{Urban Science}, 6\penalty0 (1):\penalty0 3, 2022.

\bibitem[Brinkmann et~al.(2023)Brinkmann, Swoboda, and Bartelt]{bib_gen_iccv}
Jannik Brinkmann, Paul Swoboda, and Christian Bartelt.
\newblock A multidimensional analysis of social biases in vision transformers.
\newblock In \emph{Proceedings of the IEEE/CVF International Conference on Computer Vision}, pages 4914--4923, 2023.

\bibitem[Chen et~al.(2024)Chen, Hirota, Otani, Garcia, and Nakashima]{bib_future_bias}
Tianwei Chen, Yusuke Hirota, Mayu Otani, Noa Garcia, and Yuta Nakashima.
\newblock Would deep generative models amplify bias in future models?
\newblock In \emph{Proceedings of the IEEE/CVF Conference on Computer Vision and Pattern Recognition}, pages 10833--10843, 2024.

\bibitem[Cherti et~al.(2023)Cherti, Beaumont, Wightman, Wortsman, Ilharco, Gordon, Schuhmann, Schmidt, and Jitsev]{bib_openclip}
Mehdi Cherti, Romain Beaumont, Ross Wightman, Mitchell Wortsman, Gabriel Ilharco, Cade Gordon, Christoph Schuhmann, Ludwig Schmidt, and Jenia Jitsev.
\newblock Reproducible scaling laws for contrastive language-image learning.
\newblock In \emph{Proceedings of the IEEE/CVF Conference on Computer Vision and Pattern Recognition}, pages 2818--2829, 2023.

\bibitem[Chlap et~al.(2021)Chlap, Min, Vandenberg, Dowling, Holloway, and Haworth]{bib_medical_synthetic}
Phillip Chlap, Hang Min, Nym Vandenberg, Jason Dowling, Lois Holloway, and Annette Haworth.
\newblock A review of medical image data augmentation techniques for deep learning applications.
\newblock \emph{Journal of Medical Imaging and Radiation Oncology}, 65\penalty0 (5):\penalty0 545--563, 2021.

\bibitem[Cho et~al.(2023)Cho, Zala, and Bansal]{bib_dall_eval}
Jaemin Cho, Abhay Zala, and Mohit Bansal.
\newblock Dall-eval: Probing the reasoning skills and social biases of text-to-image generation models.
\newblock In \emph{Proceedings of the IEEE/CVF International Conference on Computer Vision}, pages 3043--3054, 2023.

\bibitem[Deng et~al.(2012)Deng, Berg, Satheesh, Su, Khosla, and Li]{bib_imagenet}
Jia Deng, Alex Berg, Sanjeev Satheesh, Hao Su, Aditya Khosla, and Fei-Fei Li.
\newblock Large scale visual recognition challenge.
\newblock \emph{www. image-net. org/challenges/LSVRC/2012}, 1, 2012.

\bibitem[Dodge et~al.(2021)Dodge, Sap, Marasovi{\'c}, Agnew, Ilharco, Groeneveld, Mitchell, and Gardner]{bib_diffusion_bias_3}
Jesse Dodge, Maarten Sap, Ana Marasovi{\'c}, William Agnew, Gabriel Ilharco, Dirk Groeneveld, Margaret Mitchell, and Matt Gardner.
\newblock Documenting large webtext corpora: A case study on the colossal clean crawled corpus.
\newblock In \emph{Proceedings of the 2021 Conference on Empirical Methods in Natural Language Processing}, pages 1286--1305, 2021.

\bibitem[Elharrouss et~al.(2020)Elharrouss, Almaadeed, Al-Maadeed, and Akbari]{bib_image_inpainting}
Omar Elharrouss, Noor Almaadeed, Somaya Al-Maadeed, and Younes Akbari.
\newblock Image inpainting: A review.
\newblock \emph{Neural Processing Letters}, 51:\penalty0 2007--2028, 2020.

\bibitem[Garcia et~al.(2023)Garcia, Hirota, Wu, and Nakashima]{bib_phase}
Noa Garcia, Yusuke Hirota, Yankun Wu, and Yuta Nakashima.
\newblock Uncurated image-text datasets: Shedding light on demographic bias.
\newblock In \emph{Proceedings of the IEEE/CVF Conference on Computer Vision and Pattern Recognition}, pages 6957--6966, 2023.

\bibitem[Gustafson et~al.(2023)Gustafson, Rolland, Ravi, Duval, Adcock, Fu, Hall, and Ross]{bib_facet}
Laura Gustafson, Chloe Rolland, Nikhila Ravi, Quentin Duval, Aaron Adcock, Cheng-Yang Fu, Melissa Hall, and Candace Ross.
\newblock Facet: Fairness in computer vision evaluation benchmark.
\newblock In \emph{Proceedings of the IEEE/CVF International Conference on Computer Vision}, pages 20370--20382, 2023.

\bibitem[Hardt et~al.(2016)Hardt, Price, and Srebro]{bib_equality_of_opportunity}
Moritz Hardt, Eric Price, and Nati Srebro.
\newblock Equality of opportunity in supervised learning.
\newblock \emph{Advances in Neural Information Processing Systems}, 29, 2016.

\bibitem[Hertz et~al.()Hertz, Mokady, Tenenbaum, Aberman, Pritch, and Cohen-or]{bib_orompt_to_prompt}
Amir Hertz, Ron Mokady, Jay Tenenbaum, Kfir Aberman, Yael Pritch, and Daniel Cohen-or.
\newblock Prompt-to-prompt image editing with cross-attention control.
\newblock In \emph{The Eleventh International Conference on Learning Representations}.

\bibitem[Heusel et~al.(2017)Heusel, Ramsauer, Unterthiner, Nessler, and Hochreiter]{bib_fid}
Martin Heusel, Hubert Ramsauer, Thomas Unterthiner, Bernhard Nessler, and Sepp Hochreiter.
\newblock Gans trained by a two time-scale update rule converge to a local nash equilibrium.
\newblock \emph{Advances in neural information processing systems}, 30, 2017.

\bibitem[Hoiem et~al.(2009)Hoiem, Divvala, and Hays]{bib_pascal}
Derek Hoiem, Santosh~K Divvala, and James~H Hays.
\newblock Pascal voc 2008 challenge.
\newblock \emph{World Literature Today}, 24\penalty0 (1):\penalty0 1--4, 2009.

\bibitem[Howard et~al.(2024)Howard, Madasu, Le, Moreno, Bhiwandiwalla, and Lal]{bib_social_counterfactuals}
Phillip Howard, Avinash Madasu, Tiep Le, Gustavo~Lujan Moreno, Anahita Bhiwandiwalla, and Vasudev Lal.
\newblock Socialcounterfactuals: Probing and mitigating intersectional social biases in vision-language models with counterfactual examples.
\newblock In \emph{Proceedings of the IEEE/CVF Conference on Computer Vision and Pattern Recognition}, pages 11975--11985, 2024.

\bibitem[Jayasumana et~al.(2024)Jayasumana, Ramalingam, Veit, Glasner, Chakrabarti, and Kumar]{bib_ccmd}
Sadeep Jayasumana, Srikumar Ramalingam, Andreas Veit, Daniel Glasner, Ayan Chakrabarti, and Sanjiv Kumar.
\newblock Rethinking fid: Towards a better evaluation metric for image generation.
\newblock In \emph{Proceedings of the IEEE/CVF Conference on Computer Vision and Pattern Recognition}, pages 9307--9315, 2024.

\bibitem[Kirillov et~al.(2023)Kirillov, Mintun, Ravi, Mao, Rolland, Gustafson, Xiao, Whitehead, Berg, Lo, et~al.]{bib_sam}
Alexander Kirillov, Eric Mintun, Nikhila Ravi, Hanzi Mao, Chloe Rolland, Laura Gustafson, Tete Xiao, Spencer Whitehead, Alexander~C Berg, Wan-Yen Lo, et~al.
\newblock Segment anything.
\newblock In \emph{Proceedings of the IEEE/CVF International Conference on Computer Vision}, pages 4015--4026, 2023.

\bibitem[Le et~al.(2024)Le, Lal, and Howard]{bib_coco_counterfactuals}
Tiep Le, Vasudev Lal, and Phillip Howard.
\newblock Coco-counterfactuals: Automatically constructed counterfactual examples for image-text pairs.
\newblock \emph{Advances in Neural Information Processing Systems}, 36, 2024.

\bibitem[Lin et~al.(2014)Lin, Maire, Belongie, Hays, Perona, Ramanan, Doll{\'a}r, and Zitnick]{bib_coco}
Tsung-Yi Lin, Michael Maire, Serge Belongie, James Hays, Pietro Perona, Deva Ramanan, Piotr Doll{\'a}r, and C~Lawrence Zitnick.
\newblock Microsoft coco: Common objects in context.
\newblock In \emph{Computer Vision--ECCV 2014: 13th European Conference, Zurich, Switzerland, September 6-12, 2014, Proceedings, Part V 13}, pages 740--755. Springer, 2014.

\bibitem[Liu et~al.(2023)Liu, Zeng, Ren, Li, Zhang, Yang, Li, Yang, Su, Zhu, et~al.]{bib_grounding_dino}
Shilong Liu, Zhaoyang Zeng, Tianhe Ren, Feng Li, Hao Zhang, Jie Yang, Chunyuan Li, Jianwei Yang, Hang Su, Jun Zhu, et~al.
\newblock Grounding dino: Marrying dino with grounded pre-training for open-set object detection.
\newblock \emph{arXiv preprint arXiv:2303.05499}, 2023.

\bibitem[Luccioni et~al.(2024)Luccioni, Akiki, Mitchell, and Jernite]{bib_diffusion_bias_2}
Sasha Luccioni, Christopher Akiki, Margaret Mitchell, and Yacine Jernite.
\newblock Stable bias: Evaluating societal representations in diffusion models.
\newblock \emph{Advances in Neural Information Processing Systems}, 36, 2024.

\bibitem[Mandal et~al.(2023)Mandal, Leavy, and Little]{bib_abhishek}
Abhishek Mandal, Susan Leavy, and Suzanne Little.
\newblock Measuring bias in multimodal models: Multimodal composite association score.
\newblock In \emph{International Workshop on Algorithmic Bias in Search and Recommendation}, pages 17--30. Springer, 2023.

\bibitem[Manukyan et~al.(2023)Manukyan, Sargsyan, Atanyan, Wang, Navasardyan, and Shi]{bib_hd_painter}
Hayk Manukyan, Andranik Sargsyan, Barsegh Atanyan, Zhangyang Wang, Shant Navasardyan, and Humphrey Shi.
\newblock Hd-painter: high-resolution and prompt-faithful text-guided image inpainting with diffusion models.
\newblock \emph{arXiv preprint arXiv:2312.14091}, 2023.

\bibitem[Mishra et~al.(2022)Mishra, Panda, Phoo, Chen, Karlinsky, Saenko, Saligrama, and Feris]{bib_task2sim}
Samarth Mishra, Rameswar Panda, Cheng~Perng Phoo, Chun-Fu~Richard Chen, Leonid Karlinsky, Kate Saenko, Venkatesh Saligrama, and Rogerio~S Feris.
\newblock Task2sim: Towards effective pre-training and transfer from synthetic data.
\newblock In \emph{Proceedings of the IEEE/CVF conference on computer vision and pattern recognition}, pages 9194--9204, 2022.

\bibitem[Montalvo et~al.(2024)Montalvo, Carballeira, and Garc{\'\i}a-Mart{\'\i}n]{bib_montalvo}
Javier Montalvo, Pablo Carballeira, and {\'A}lvaro Garc{\'\i}a-Mart{\'\i}n.
\newblock Synthmanticlidar: A synthetic dataset for semantic segmentation on lidar imaging.
\newblock In \emph{2024 IEEE International Conference on Image Processing (ICIP)}, pages 137--143. IEEE, 2024.

\bibitem[Moreu et~al.(2023{\natexlab{a}})Moreu, Arazo, McGuinness, and O'Connor]{bib_noel1}
Enric Moreu, Eric Arazo, Kevin McGuinness, and Noel~E O'Connor.
\newblock Joint one-sided synthetic unpaired image translation and segmentation for colorectal cancer prevention.
\newblock \emph{Expert Systems}, 40\penalty0 (6):\penalty0 e13137, 2023{\natexlab{a}}.

\bibitem[Moreu et~al.(2023{\natexlab{b}})Moreu, Arazo, McGuinness, and O'Connor]{bib_noel2}
Enric Moreu, Eric Arazo, Kevin McGuinness, and Noel~E O'Connor.
\newblock Self-supervised and semi-supervised polyp segmentation using synthetic data.
\newblock In \emph{2023 International Joint Conference on Neural Networks (IJCNN)}, pages 1--9. IEEE, 2023{\natexlab{b}}.

\bibitem[Ramaswamy et~al.(2021)Ramaswamy, Kim, and Russakovsky]{bib_dataset_biases_2}
Vikram~V Ramaswamy, Sunnie~SY Kim, and Olga Russakovsky.
\newblock Fair attribute classification through latent space de-biasing.
\newblock In \emph{Proceedings of the IEEE/CVF Conference on Computer Vision and Pattern Recognition}, pages 9301--9310, 2021.

\bibitem[Ramesh et~al.(2021)Ramesh, Pavlov, Goh, Gray, Voss, Radford, Chen, and Sutskever]{bib_dalle}
Aditya Ramesh, Mikhail Pavlov, Gabriel Goh, Scott Gray, Chelsea Voss, Alec Radford, Mark Chen, and Ilya Sutskever.
\newblock Zero-shot text-to-image generation.
\newblock In \emph{International conference on machine learning}, pages 8821--8831. Pmlr, 2021.

\bibitem[Ravi et~al.(2024)Ravi, Gabeur, Hu, Hu, Ryali, Ma, Khedr, R{\"a}dle, Rolland, Gustafson, et~al.]{bib_sam2}
Nikhila Ravi, Valentin Gabeur, Yuan-Ting Hu, Ronghang Hu, Chaitanya Ryali, Tengyu Ma, Haitham Khedr, Roman R{\"a}dle, Chloe Rolland, Laura Gustafson, et~al.
\newblock Sam 2: Segment anything in images and videos.
\newblock \emph{arXiv preprint arXiv:2408.00714}, 2024.

\bibitem[Ricanek and Tesafaye(2006)]{bib_morph}
Karl Ricanek and Tamirat Tesafaye.
\newblock Morph: A longitudinal image database of normal adult age-progression.
\newblock In \emph{7th international conference on automatic face and gesture recognition (FGR06)}, pages 341--345. IEEE, 2006.

\bibitem[Richter et~al.(2016)Richter, Vineet, Roth, and Koltun]{bib_syn3}
Stephan~R Richter, Vibhav Vineet, Stefan Roth, and Vladlen Koltun.
\newblock Playing for data: Ground truth from computer games.
\newblock In \emph{Computer Vision--ECCV 2016: 14th European Conference, Amsterdam, The Netherlands, October 11-14, 2016, Proceedings, Part II 14}, pages 102--118. Springer, 2016.

\bibitem[Rombach et~al.(2022)Rombach, Blattmann, Lorenz, Esser, and Ommer]{bib_stable_diffusion}
Robin Rombach, Andreas Blattmann, Dominik Lorenz, Patrick Esser, and Bj{\"o}rn Ommer.
\newblock High-resolution image synthesis with latent diffusion models.
\newblock In \emph{Proceedings of the IEEE/CVF conference on computer vision and pattern recognition}, pages 10684--10695, 2022.

\bibitem[Saharia et~al.(2022)Saharia, Chan, Saxena, Li, Whang, Denton, Ghasemipour, Gontijo~Lopes, Karagol~Ayan, Salimans, et~al.]{bib_imagen}
Chitwan Saharia, William Chan, Saurabh Saxena, Lala Li, Jay Whang, Emily~L Denton, Kamyar Ghasemipour, Raphael Gontijo~Lopes, Burcu Karagol~Ayan, Tim Salimans, et~al.
\newblock Photorealistic text-to-image diffusion models with deep language understanding.
\newblock \emph{Advances in neural information processing systems}, 35:\penalty0 36479--36494, 2022.

\bibitem[Schuhmann et~al.(2022{\natexlab{a}})Schuhmann, Beaumont, Vencu, Gordon, Wightman, Cherti, Coombes, Katta, Mullis, Wortsman, et~al.]{bib_ar}
Christoph Schuhmann, Romain Beaumont, Richard Vencu, Cade Gordon, Ross Wightman, Mehdi Cherti, Theo Coombes, Aarush Katta, Clayton Mullis, Mitchell Wortsman, et~al.
\newblock Laion-5b: An open large-scale dataset for training next generation image-text models.
\newblock \emph{Advances in Neural Information Processing Systems}, 35:\penalty0 25278--25294, 2022{\natexlab{a}}.

\bibitem[Schuhmann et~al.(2022{\natexlab{b}})Schuhmann, Beaumont, Vencu, Gordon, Wightman, Cherti, Coombes, Katta, Mullis, Wortsman, et~al.]{bib_laion5b}
Christoph Schuhmann, Romain Beaumont, Richard Vencu, Cade Gordon, Ross Wightman, Mehdi Cherti, Theo Coombes, Aarush Katta, Clayton Mullis, Mitchell Wortsman, et~al.
\newblock Laion-5b: An open large-scale dataset for training next generation image-text models.
\newblock \emph{Advances in Neural Information Processing Systems}, 35:\penalty0 25278--25294, 2022{\natexlab{b}}.

\bibitem[Sharma et~al.(2018)Sharma, Ding, Goodman, and Soricut]{bib_cc3m}
Piyush Sharma, Nan Ding, Sebastian Goodman, and Radu Soricut.
\newblock Conceptual captions: A cleaned, hypernymed, image alt-text dataset for automatic image captioning.
\newblock In \emph{Proceedings of the 56th Annual Meeting of the Association for Computational Linguistics (Volume 1: Long Papers)}, pages 2556--2565, 2018.

\bibitem[Shumailov et~al.(2024)Shumailov, Shumaylov, Zhao, Papernot, Anderson, and Gal]{shumailov2024ai}
Ilia Shumailov, Zakhar Shumaylov, Yiren Zhao, Nicolas Papernot, Ross Anderson, and Yarin Gal.
\newblock Ai models collapse when trained on recursively generated data.
\newblock \emph{Nature}, 631\penalty0 (8022):\penalty0 755--759, 2024.

\bibitem[Sirotkin et~al.(2022)Sirotkin, Carballeira, and Escudero-Vi\~nolo]{bib_sirotkin}
Kirill Sirotkin, Pablo Carballeira, and Marcos Escudero-Vi\~nolo.
\newblock A study on the distribution of social biases in self-supervised learning visual models.
\newblock In \emph{Proceedings of the IEEE/CVF Conference on Computer Vision and Pattern Recognition}, pages 10442--10451, 2022.

\bibitem[Siu et~al.(2022)Siu, Wang, and Cheng]{bib_pipe}
ChunFai Siu, Mingzhu Wang, and Jack~CP Cheng.
\newblock A framework for synthetic image generation and augmentation for improving automatic sewer pipe defect detection.
\newblock \emph{Automation in Construction}, 137:\penalty0 104213, 2022.

\bibitem[Steed and Caliskan(2021)]{bib_biases_ieat}
Ryan Steed and Aylin Caliskan.
\newblock Image representations learned with unsupervised pre-training contain human-like biases.
\newblock In \emph{Proceedings of the 2021 ACM Conference on Fairness, Accountability, and Transparency}, pages 701--713, 2021.

\bibitem[Struppek et~al.(2023)Struppek, Hintersdorf, Friedrich, Schramowski, Kersting, et~al.]{bib_gen_fail}
Lukas Struppek, Dom Hintersdorf, Felix Friedrich, Patrick Schramowski, Kristian Kersting, et~al.
\newblock Exploiting cultural biases via homoglyphs in text-to-image synthesis.
\newblock \emph{Journal of Artificial Intelligence Research}, 78:\penalty0 1017--1068, 2023.

\bibitem[Su et~al.(2015)Su, Qi, Li, and Guibas]{bib_syn4}
Hao Su, Charles~R Qi, Yangyan Li, and Leonidas~J Guibas.
\newblock Render for cnn: Viewpoint estimation in images using cnns trained with rendered 3d model views.
\newblock In \emph{Proceedings of the IEEE international conference on computer vision}, pages 2686--2694, 2015.

\bibitem[Thong et~al.(2023)Thong, Joniak, and Xiang]{bib_beyond_skin_tone}
William Thong, Przemyslaw Joniak, and Alice Xiang.
\newblock Beyond skin tone: A multidimensional measure of apparent skin color.
\newblock In \emph{Proceedings of the IEEE/CVF International Conference on Computer Vision}, pages 4903--4913, 2023.

\bibitem[Tobin et~al.(2017)Tobin, Fong, Ray, Schneider, Zaremba, and Abbeel]{bib_syn5}
Josh Tobin, Rachel Fong, Alex Ray, Jonas Schneider, Wojciech Zaremba, and Pieter Abbeel.
\newblock Domain randomization for transferring deep neural networks from simulation to the real world.
\newblock In \emph{2017 IEEE/RSJ international conference on intelligent robots and systems (IROS)}, pages 23--30. IEEE, 2017.

\bibitem[Wang and Russakovsky(2023)]{bib_overwriting_biases}
Angelina Wang and Olga Russakovsky.
\newblock Overwriting pretrained bias with finetuning data.
\newblock In \emph{Proceedings of the IEEE/CVF International Conference on Computer Vision}, pages 3957--3968, 2023.

\bibitem[Wang et~al.(2022)Wang, Barocas, Laird, and Wallach]{bib_dataset_biases_1}
Angelina Wang, Solon Barocas, Kristen Laird, and Hanna Wallach.
\newblock Measuring representational harms in image captioning.
\newblock In \emph{Proceedings of the 2022 ACM Conference on Fairness, Accountability, and Transparency}, pages 324--335, 2022.

\bibitem[Wang et~al.(2019)Wang, Zhao, Yatskar, Chang, and Ordonez]{bib_balanced_datasets}
Tianlu Wang, Jieyu Zhao, Mark Yatskar, Kai-Wei Chang, and Vicente Ordonez.
\newblock Balanced datasets are not enough: Estimating and mitigating gender bias in deep image representations.
\newblock In \emph{Proceedings of the IEEE/CVF International Conference on Computer Vision}, pages 5310--5319, 2019.

\bibitem[Wang et~al.(2024)Wang, Li, Luo, Wang, and De~la Torre]{bib_model_diagnosis}
Yinong~Oliver Wang, Eileen Li, Jinqi Luo, Zhaoning Wang, and Fernando De~la Torre.
\newblock Unsupervised model diagnosis.
\newblock \emph{arXiv preprint arXiv:2410.06243}, 2024.

\bibitem[Wei et~al.(2023)Wei, Hu, Xie, Liu, Zhang, Cao, Bao, Chen, and Guo]{bib_finetuning_clip}
Yixuan Wei, Han Hu, Zhenda Xie, Ze Liu, Zheng Zhang, Yue Cao, Jianmin Bao, Dong Chen, and Baining Guo.
\newblock Improving clip fine-tuning performance.
\newblock In \emph{Proceedings of the IEEE/CVF International Conference on Computer Vision}, pages 5439--5449, 2023.

\bibitem[Wolfe and Caliskan(2022)]{bib_markedness}
Robert Wolfe and Aylin Caliskan.
\newblock Markedness in visual semantic ai.
\newblock In \emph{Proceedings of the 2022 ACM Conference on Fairness, Accountability, and Transparency}, pages 1269--1279, 2022.

\bibitem[Wolfe et~al.(2023)Wolfe, Yang, Howe, and Caliskan]{bib_biases_4}
Robert Wolfe, Yiwei Yang, Bill Howe, and Aylin Caliskan.
\newblock Contrastive language-vision ai models pretrained on web-scraped multimodal data exhibit sexual objectification bias.
\newblock In \emph{Proceedings of the 2023 ACM Conference on Fairness, Accountability, and Transparency}, pages 1174--1185, 2023.

\bibitem[Wu et~al.(2023)Wu, Sun, Zhu, Zhao, and Li]{bib_hps}
Xiaoshi Wu, Keqiang Sun, Feng Zhu, Rui Zhao, and Hongsheng Li.
\newblock Human preference score: Better aligning text-to-image models with human preference.
\newblock In \emph{Proceedings of the IEEE/CVF International Conference on Computer Vision}, pages 2096--2105, 2023.

\bibitem[Wu et~al.(2024)Wu, Nakashima, and Garcia]{bib_diffusion_bias_1}
Yankun Wu, Yuta Nakashima, and Noa Garcia.
\newblock Stable diffusion exposed: Gender bias from prompt to image.
\newblock In \emph{Proceedings of the AAAI/ACM Conference on AI, Ethics, and Society}, pages 1648--1659, 2024.

\bibitem[Xu et~al.(2024)Xu, Liu, Wu, Tong, Li, Ding, Tang, and Dong]{bib_ir}
Jiazheng Xu, Xiao Liu, Yuchen Wu, Yuxuan Tong, Qinkai Li, Ming Ding, Jie Tang, and Yuxiao Dong.
\newblock Imagereward: Learning and evaluating human preferences for text-to-image generation.
\newblock \emph{Advances in Neural Information Processing Systems}, 36, 2024.

\bibitem[Xuan et~al.(2024)Xuan, Xian, Xintao, Yuxuan, Ying, and Qiang]{bib_brushnet}
JU Xuan, Liu Xian, Wang Xintao, Bian Yuxuan, Shan Ying, and Xu Qiang.
\newblock Brushnet: A plug-and-play image inpainting model with decomposed dual-branch diffusion.
\newblock In \emph{Proceedings of the European Conference on Computer Vision}, 2024.

\bibitem[Zhang et~al.(2023{\natexlab{a}})Zhang, Rao, and Agrawala]{bib_control_net}
Lvmin Zhang, Anyi Rao, and Maneesh Agrawala.
\newblock Adding conditional control to text-to-image diffusion models.
\newblock In \emph{Proceedings of the IEEE/CVF International Conference on Computer Vision}, pages 3836--3847, 2023{\natexlab{a}}.

\bibitem[Zhang et~al.(2023{\natexlab{b}})Zhang, Jiang, Turk, and Yang]{bib_gender_representation}
Yanzhe Zhang, Lu Jiang, Greg Turk, and Diyi Yang.
\newblock Auditing gender presentation differences in text-to-image models.
\newblock \emph{arXiv preprint arXiv:2302.03675}, 2023{\natexlab{b}}.

\bibitem[Zhao et~al.(2017)Zhao, Wang, Yatskar, Ordonez, and Chang]{bib_men_shopping}
Jieyu Zhao, Tianlu Wang, Mark Yatskar, Vicente Ordonez, and Kai-Wei Chang.
\newblock Men also like shopping: Reducing gender bias amplification using corpus-level constraints.
\newblock In \emph{Proceedings of the 2017 Conference on Empirical Methods in Natural Language Processing}, pages 2979--2989, Copenhagen, Denmark, 2017. Association for Computational Linguistics.

\bibitem[Zhuang et~al.(2023)Zhuang, Zeng, Liu, Yuan, and Chen]{bib_power_paint}
Junhao Zhuang, Yanhong Zeng, Wenran Liu, Chun Yuan, and Kai Chen.
\newblock A task is worth one word: Learning with task prompts for high-quality versatile image inpainting.
\newblock \emph{arXiv preprint arXiv:2312.03594}, 2023.

\end{thebibliography}
}

\clearpage
\setcounter{page}{1}
\maketitlesupplementary

\section{Visual examples of Pinpoint counterfactuals}
\label{sec_more_examples}

\begin{figure*}[ht!]
  \centering
\begin{subfigure}[l]{0.19\textwidth}
\centering
      \includegraphics[height=5cm]{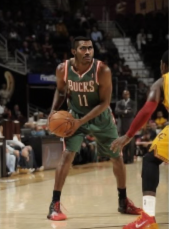}
 \end{subfigure}
  \begin{subfigure}[l]{0.19\textwidth}
  \centering
      \includegraphics[height=5cm]{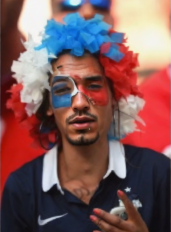}
 \end{subfigure}
  \begin{subfigure}[l]{0.19\textwidth}
  \centering
      \includegraphics[height=5cm]{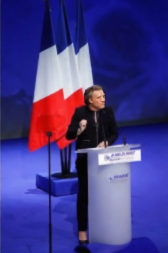}
 \end{subfigure}
  \begin{subfigure}[l]{0.19\textwidth}
  \centering
      \includegraphics[height=5cm]{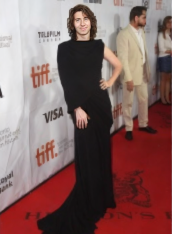}
 \end{subfigure}
  \begin{subfigure}[l]{0.19\textwidth}
  \centering
      \includegraphics[height=5cm]{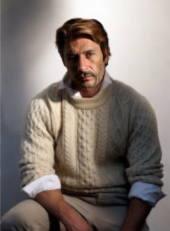}
 \end{subfigure}
 \vfill
 \begin{subfigure}[l]{0.19\textwidth}
\centering
      \includegraphics[height=5cm]{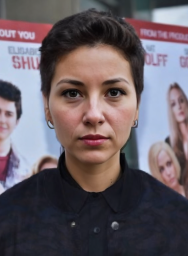}
 \end{subfigure}
  \begin{subfigure}[l]{0.19\textwidth}
  \centering
      \includegraphics[height=5cm]{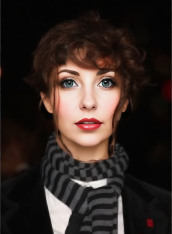}
 \end{subfigure}
  \begin{subfigure}[l]{0.19\textwidth}
  \centering
      \includegraphics[height=5cm]{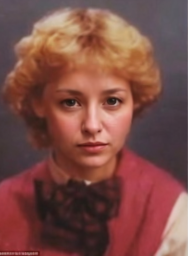}
 \end{subfigure}
  \begin{subfigure}[l]{0.19\textwidth}
  \centering
      \includegraphics[height=5cm]{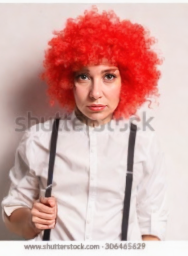}
 \end{subfigure}
  \begin{subfigure}[l]{0.19\textwidth}
  \centering
      \includegraphics[height=5cm]{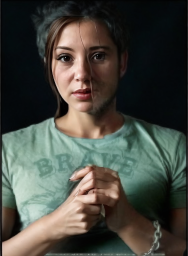}
 \end{subfigure}

  \caption{Examples of in-painted images from CC3M dataset \cite{bib_cc3m}. Top row: in-painted images of men, bottom row: in-painted images of women.}
  \label{fig_examples_supp}
  \vspace{-0.3cm}
\end{figure*}

\begin{figure*}[ht!]
  \centering
  \begin{subfigure}[l]{0.23\textwidth}
  \centering
      \includegraphics[height=5cm]{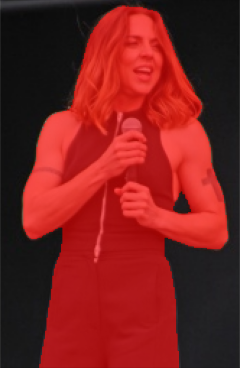}
 \end{subfigure}
  \begin{subfigure}[l]{0.23\textwidth}
  \centering
      \includegraphics[height=5cm]{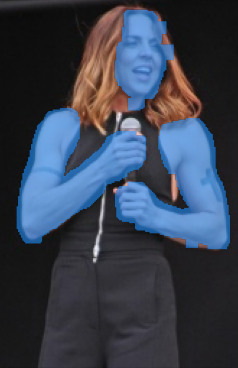}
 \end{subfigure}
  \begin{subfigure}[l]{0.23\textwidth}
  \centering
      \includegraphics[height=5cm]{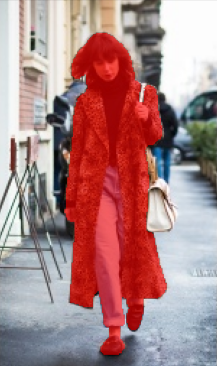}
 \end{subfigure}
  \begin{subfigure}[l]{0.23\textwidth}
  \centering
      \includegraphics[height=5cm]{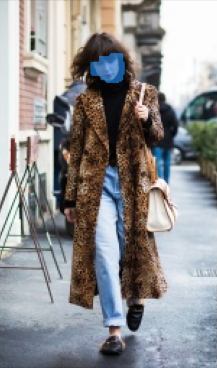}
 \end{subfigure}
  \vfill
  \begin{subfigure}[l]{0.49\textwidth}
  \centering
      \includegraphics[height=5cm]{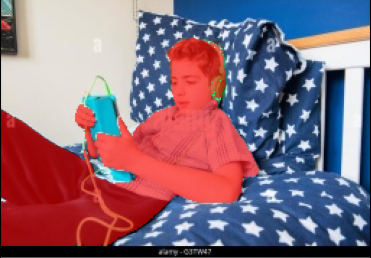}
 \end{subfigure}
  \begin{subfigure}[l]{0.49\textwidth}
  \centering
      \includegraphics[height=5cm]{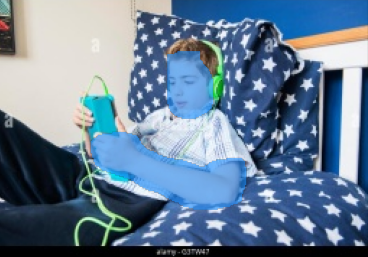}
 \end{subfigure}

  \caption{Examples of mask generated with Lang-SAM for the attribute ``skin'' (shown in red), and our method (shown in blue).}
  \label{fig_masks}
  \vspace{-0.3cm}
\end{figure*}

In this section we provide additional examples of in-panted images generated using Pinpoint counterfactual in-painting approach (Figure \ref{fig_examples_supp}), and comparison between the masks generated using our method and Language-SAM~\footnote{An open-source project combining Segment Anything Model 2 \cite{bib_sam2} and GroundingDINO \cite{bib_grounding_dino}, accessible at \url{https://github.com/luca-medeiros/lang-segment-anything}} (Figure \ref{fig_masks}). We use the keyword ``skin'' as a prompt for the Language-SAM model, and as can be seen from Figure \ref{fig_masks}, the model is not able to distinguish such fine-grain details, instead producing a mask for the entire person. On the contrary, our approach, produces masks for the desired image regions.

\section{Equality of opportunity}

As mentioned in Section 5.4 of the main paper, we compute recall on the selected occupations of FACET \cite{bib_facet}, measuring the equality of opportunity for the two genders. The original dataset contains annotations for 52 different occupations. To select a list of occupations used in this work, we follow the original methodology \cite{bib_facet}, and only select images that contain a single annotated person. Then, to obtain statistically significant results, we select the occupations with more than 50 examples of both genders. Finally, we filter out the occupations that do not appear in the in-painted images of CC3M \footnote{Since CC3M does not have ground-truth annotations, we search for occupations using their names as a keyword in a caption describing each image.}, since we don't train the models for the downstream task (the inference is done in zero-shot regime). Following this approach, we are left with the occupations summarized in Table \ref{table_facet_summary}. The complete data on per-gender recall used in Section 5.4 and Figure 4, is presented in Table \ref{table_recalls}.

\begin{table}[]
    \centering
    \begin{tabular}{@{}cccc@{}}
\toprule
Occupation &
  \begin{tabular}[c]{@{}c@{}}\# males \\ in FACET\end{tabular} &
  \begin{tabular}[c]{@{}c@{}}\# females\\  in FACET\end{tabular} &
  \begin{tabular}[c]{@{}c@{}}\# appearances\\  in CC3M\end{tabular} \\ \midrule
Dancer  & 132 & 156 & 517 \\
Doctor  & 95  & 99  & 853 \\
Farmer  & 488 & 255 & 309 \\
Nurse   & 70  & 158 & 388 \\
Patient & 131 & 72  & 483 \\
Singer  & 507 & 251 & 951 \\ \bottomrule
\end{tabular}
    \caption{Summary of the occupations used in the experiments that measure equality of opportunity.}
    \label{table_facet_summary}
\end{table}

\section{Does fine-tuning on in-painted data balanced for one attribute affect performance on other attributes?}

We address this question using zero-shot image retrieval on PHASE \cite{bib_phase} dataset and attributes age (young, adult, senior) and skin-tone (lighter, darker). Table \ref{table_image_retrieval} shows that the image retrieval performance for models finetuned on gender-balanced data (c4-c7), in which one of the gender fully consists of synthetic counterfactual images, does not change significantly relative to the model trained on the real data (c1-c3). This shows that the age and skin-tone distributions of people on in-painted synthetic images is close to the original images and does not negatively affect model's performance for these attributes. %Furthermore, while the average values of self-similarity do not change significantly for the models finetuned with or without synthetic data (see Table \ref{table_finetuning}).

\section{Automated captions editing}

Given an original caption $C_O$, our goal is to produce a new caption $C_{A_i}$ for each counterfactual sample $i$ that we generate, where $A$ is the value of a protected attribute that is being in-painted in the image. In this work, we focus on the gender counterfactuals, therefore, for each original caption $C_O$ we produce a pair of captions $C_M$ and $C_W$ that correspond to the two genders.

To determine if $C_O$ contains descriptions of a gender, we search for keywords that explicitly or implicitly indicate a specific gender. For example, words like \textit{'he', 'man', 'men', 'male', 'mister', ...} explicitly reveal that the gender is masculine, while occupation/activity-related words like \textit{'paperboy', 'policeman', 'businessman', 'waiter', 'actor'}, do so implicitly (the contrary is true for women). We, thus, construct two sets of gender-indicating keywords: $K_M$ and $K_W$ and establish correspondences between them $f_1:K_M\rightarrow K_W$ and $f_2:K_W\rightarrow K_M$ (i.e., \textit{man} $\rightarrow$ \textit{woman}, \textit{policeman} $\rightarrow$ \textit{policewoman}, \textit{actor} $\rightarrow$ \textit{actress}, \textit{queen} $\rightarrow$ \textit{king}, \textit{female} $\rightarrow$ \textit{male}, etc.).

In the case if $C_O$ does not indicate the gender of (a) person(s), the caption is left unchanged, i.e. $C_M = C_W = C_O$. In the case if $C_O$ describes a man (men), $C_O = C_M$ and $C_W$ is generated, otherwise $C_O = C_W$ and $C_M$ is generated. Then, for each word $w$ in $C_O$, if $w \in K_M$ and the target gender is "woman", then we substitute $w$ with $\hat{w} = f_1(w)$. Alternatively, if $w \in K_W$ and the target gender is "man", then we substitute $w$ with $\hat{w} = f_2(w)$. We provide examples of edited captions in Table \ref{table_captions}.

% \usepackage{booktabs}
% \usepackage[normalem]{ulem}

\begin{table*}[]
\centering
\begin{tabular}{@{}c|c@{}}
\toprule
Original  & \textbf{man} buying some fruit on the market , selective focus   \\
Masculine & \textbf{man} buying some fruit on the market , selective focus   \\
Feminine  & \textbf{woman} buying some fruit on the market , selective focus \\ \midrule
Original  & \textbf{actor} in garment with artist                            \\
Masculine & \textbf{actor} in garment with artist                            \\
Feminine  & \textbf{actress} in garment with artist                          \\ \midrule
Original  & painting of a young \textbf{woman} dressed as video game series  \\
Masculine & painting of a young \textbf{man} dressed as video game series    \\
Feminine  & painting of a young \textbf{woman} dressed as video game series  \\ \midrule
Original  & \textbf{actress} with a beautiful smile                          \\
Masculine & \textbf{actor} with a beautiful smile                            \\
Feminine  & \textbf{actress} with a beautiful smile                          \\ \midrule
Original  & \textbf{person} , was surprised by the staff                     \\
Masculine & \textbf{person} , was surprised by the staff                     \\
Feminine  & \textbf{person} , was surprised by the staff                     \\ \bottomrule
\end{tabular}
\caption{Examples of automated captions editing. First two top captions: original gender is masculine, third and fourth captions: original gender is feminine, last caption: original gender is unknown.}
\label{table_captions}
\end{table*}

\begin{table*}[]
\centering
\begin{tabular}{@{}cccccccccc@{}}
\toprule
Data type &
  \begin{tabular}[c]{@{}c@{}}Finetuning\\ Dataset\end{tabular} &
  All @1 &
  \begin{tabular}[c]{@{}c@{}}M\\ \#1950\end{tabular} &
  \begin{tabular}[c]{@{}c@{}}W\\ \#1617\end{tabular} &
  \begin{tabular}[c]{@{}c@{}}Young\\ \#1349\end{tabular} &
  \begin{tabular}[c]{@{}c@{}}Adult\\ \#1509\end{tabular} &
  \begin{tabular}[c]{@{}c@{}}Senior\\ \#128\end{tabular} &
  \begin{tabular}[c]{@{}c@{}}Lighter\\ \#3166\end{tabular} &
  \begin{tabular}[c]{@{}c@{}}Darker\\ \#318\end{tabular} \\ \midrule
\multicolumn{1}{c|}{} &
  None &
  37.62 &
  \cellcolor[HTML]{EFEFEF}36.15 &
  \cellcolor[HTML]{EFEFEF}39.70 &
  37.58 &
  31.68 &
  {\color[HTML]{CB0000} 47.66} &
  37.30 &
  {\color[HTML]{CB0000} 34.59} \\
\multicolumn{1}{c|}{} &
  c1 &
  40.37 &
  \cellcolor[HTML]{EFEFEF}39.74 &
  \cellcolor[HTML]{EFEFEF}41.80 &
  39.73 &
  {\color[HTML]{036400} 34.66} &
  {\color[HTML]{036400} 53.91} &
  39.99 &
  38.68 \\
\multicolumn{1}{c|}{} &
  c2 &
  {\color[HTML]{036400} 40.75} &
  \cellcolor[HTML]{EFEFEF}39.23 &
  \cellcolor[HTML]{EFEFEF}{\color[HTML]{036400} 43.29} &
  39.96 &
  34.53 &
  51.57 &
  40.05 &
  38.99 \\
\multicolumn{1}{c|}{\multirow{-4}{*}{Real}} &
  c3 &
  {\color[HTML]{000000} 41.57} &
  \cellcolor[HTML]{EFEFEF}{\color[HTML]{036400} 40.77} &
  \cellcolor[HTML]{EFEFEF}{\color[HTML]{000000} 42.98} &
  {\color[HTML]{036400} 40.99} &
  {\color[HTML]{036400} 35.52} &
  {\color[HTML]{036400} 53.91} &
  {\color[HTML]{000000} 40.75} &
  {\color[HTML]{000000} 41.19} \\ \midrule
\multicolumn{1}{c|}{} &
  c4 &
  40.51 &
  \cellcolor[HTML]{EFEFEF}39.08 &
  \cellcolor[HTML]{EFEFEF}42.55 &
  {\color[HTML]{000000} 40.40} &
  34.38 &
  {\color[HTML]{036400} 53.91} &
  {\color[HTML]{000000} 40.18} &
  38.68 \\
\multicolumn{1}{c|}{} &
  c5 &
  40.79 &
  \cellcolor[HTML]{EFEFEF}40.21 &
  \cellcolor[HTML]{EFEFEF}41.81 &
  40.85 &
  33.66 &
  53.13 &
  39.83 &
  41.82 \\
\multicolumn{1}{c|}{} &
  c6 &
  40.64 &
  \cellcolor[HTML]{EFEFEF}{\color[HTML]{000000} 40.56} &
  \cellcolor[HTML]{EFEFEF}40.57 &
  39.58 &
  {\color[HTML]{000000} 34.53} &
  {\color[HTML]{036400} 53.91} &
  39.80 &
  {\color[HTML]{036400} 41.82} \\
\multicolumn{1}{c|}{} &
  c7 &
  41.48 &
  \cellcolor[HTML]{EFEFEF}40.56 &
  \cellcolor[HTML]{EFEFEF}42.24 &
  40.70 &
  35.12 &
  54.69 &
  {\color[HTML]{036400} 40.78} &
  42.46 \\
\multicolumn{1}{c|}{\multirow{-5}{*}{\begin{tabular}[c]{@{}c@{}}Real + \\ ours\end{tabular}}} &
  c8 &
  39.97 &
  \cellcolor[HTML]{EFEFEF}39.54 &
  \cellcolor[HTML]{EFEFEF}40.88 &
  39.51 &
  34.53 &
  52.34 &
  39.86 &
  40.57 \\ \midrule
\multicolumn{1}{c|}{} &
  c9 &
  {\color[HTML]{CB0000} 36.99} &
  \cellcolor[HTML]{EFEFEF}{\color[HTML]{CB0000} \textbf{36.56}} &
  \cellcolor[HTML]{EFEFEF}{\color[HTML]{CB0000} \textbf{37.11}} &
  {\color[HTML]{CB0000} 35.66} &
  {\color[HTML]{CB0000} 31.28} &
  52.32 &
  {\color[HTML]{CB0000} 35.91} &
  36.16 \\
\multicolumn{1}{c|}{\multirow{-2}{*}{Ours}} &
  c10 &
  40.18 &
  \cellcolor[HTML]{EFEFEF}39.85 &
  \cellcolor[HTML]{EFEFEF}40.94 &
  39.36 &
  33.53 &
  57.81 &
  39.23 &
  40.25 \\ \bottomrule
\end{tabular}
\caption{Image retrieval scores on PHASE \cite{bib_phase} dataset for a ViT-B/16 CLIP model pretrained on LAION-2B \cite{bib_laion5b} and finetuned on real or real + in-painted data generated using the proposed framework.}
\label{table_image_retrieval}
\end{table*}

\begin{table*}[]
\begin{tabular}{@{}ccccccccccccc@{}}
\toprule
\multirow{2}{*}{\begin{tabular}[c]{@{}c@{}}Finetuning\\ Dataset\end{tabular}} &
  \multicolumn{2}{c}{Dancer} &
  \multicolumn{2}{c}{Doctor} &
  \multicolumn{2}{c}{Patient} &
  \multicolumn{2}{c}{Farmer} &
  \multicolumn{2}{c}{Nurse} &
  \multicolumn{2}{c}{Singer} \\ \cmidrule(l){2-13} 
     & M     & W     & M     & W     & M     & W     & M     & W     & M     & W     & M     & W     \\ \midrule
None & 50.70 & 58.43 & 28.12 & 22.00 & 84.56 & 90.91 & 78.12 & 74.22 & 12.50 & 20.00 & 76.13 & 80.08 \\
c1   & 53.52 & 53.01 & 19.79 & 17.00 & 88.24 & 89.61 & 76.69 & 68.75 & 15.28 & 20.62 & 80.67 & 88.45 \\
c2   & 50.00 & 50.60 & 29.17 & 22.00 & 86.03 & 90.91 & 77.91 & 69.53 & 6.94  & 11.25 & 79.88 & 85.26 \\
c3   & 54.93 & 54.82 & 20.83 & 13.00 & 83.82 & 89.61 & 75.66 & 67.19 & 15.28 & 25.00 & 81.07 & 87.65 \\ \midrule
c4   & 43.66 & 47.59 & 22.92 & 23.00 & 85.29 & 88.31 & 80.37 & 74.61 & 2.78  & 5.63  & 75.74 & 83.27 \\
c5   & 51.41 & 53.61 & 19.79 & 21.00 & 83.82 & 89.61 & 76.07 & 67.58 & 9.72  & 17.50 & 82.25 & 88.45 \\
c6   & 61.27 & 56.02 & 11.46 & 8.00  & 84.56 & 88.31 & 75.66 & 67.19 & 30.56 & 25.62 & 87.77 & 88.84 \\
c7   & 52.11 & 51.81 & 16.67 & 14.00 & 84.56 & 90.91 & 77.30 & 70.70 & 12.50 & 19.38 & 80.67 & 86.85 \\
c8   & 61.97 & 69.88 & 20.83 & 16.00 & 47.79 & 38.96 & 86.71 & 82.03 & 56.94 & 68.75 & 60.55 & 61.75 \\ \midrule
c9   & 57.04 & 49.40 & 14.58 & 14.00 & 83.09 & 83.12 & 72.60 & 65.23 & 19.44 & 17.50 & 87.57 & 87.25 \\
c10  & 52.11 & 52.41 & 19.79 & 13.00 & 80.88 & 83.12 & 76.89 & 68.75 & 11.11 & 21.88 & 81.46 & 87.25 \\ \bottomrule
\end{tabular}
\caption{Per-gender (M, W) recall on select occupations of FACET dataset \cite{bib_facet}. The inference is done in zero-shot mode with ViT-B/16 CLIP pretrained on LAION-2B \cite{bib_laion5b} and fine-tuned using the combinations of real and synthetic data described in Table 2 of the main paper.}
\label{table_recalls}
\end{table*}

\end{document}